\pdfoutput=1

\documentclass[11pt]{article}
\usepackage{setspace}
\usepackage{multirow}
\usepackage[dvipsnames,table,xcdraw]{xcolor}

\usepackage{graphicx}
\usepackage{tabularx}
\usepackage{afterpage}
\usepackage{booktabs}
\usepackage{amssymb}
\usepackage{amsmath}
\usepackage{makecell}

\usepackage{subcaption}
\usepackage{anyfontsize}

\usepackage[]{acl}

\usepackage{times}
\usepackage{latexsym}
\usepackage{xspace}
\usepackage{tcolorbox}

\usepackage[T1]{fontenc}

\usepackage[utf8]{inputenc}

\usepackage{microtype}

\usepackage{inconsolata}

\usepackage{graphicx}
\usepackage{cleveref}
\usepackage[normalem]{ulem}

\title{Conversations Gone Awry, But Then?\\ Evaluating Conversational Forecasting Models}

\author{
  \textbf{Son Quoc Tran} \quad \textbf{Tushaar Gangavarapu} \quad \textbf{ Nicholas Chernogor} \\ \textbf{Jonathan P. Chang} \quad \textbf{Cristian Danescu-Niculescu-Mizil} \\
  Cornell University \quad The University of Texas at Austin \quad Harvey Mudd College \\
  \texttt{\{sontran, cristian\}@cs.cornell.edu}\\
  \texttt{tushaarg@cs.utexas.edu} \quad \texttt{ nac86@cornell.edu} \quad \texttt{jpchang@g.hmc.edu} 
}

\newif\ifshowcomments

\newcommand{\cut}[1]{}
\newcommand{\xhdr}[1]{\paragraph{#1.}}

\newcommand{\forecastingmodel}{conversational forecasting model\xspace}

\newcommand{\wiki}{CGA-Wiki\xspace}
\newcommand{\cgacmv}{CGA-CMV\xspace}
\newcommand{\oldcmv}{CGA-CMV-legacy\xspace}
\newcommand{\newcmv}{CGA-CMV-large\xspace}

\newcommand{\indiforecast}{\widehat{y}_t\xspace}

\newcommand{\binarylastforecast}{\widehat{g_{n-1}}\xspace}

\newcommand{\incorrcount}{IR}
\newcommand{\corrcount}{CR}

\newcommand{\recovery}{recovery\xspace}
\newcommand{\Recovery}{Recovery\xspace}
\newcommand{\recoveries}{recoveries\xspace}

\newcommand{\furl}[1]{\footnote{\url{#1}.}}

\begin{document}
\maketitle

\begin{abstract}

We often rely on our intuition to anticipate the direction of a conversation.
Endowing automated systems with similar foresight can enable them to assist human-human interactions.
Recent work on developing models with this predictive capacity has focused on the Conversations Gone Awry (CGA) task: forecasting whether an ongoing conversation will derail.
In this work, we revisit this task and introduce the first uniform evaluation framework, creating a benchmark that enables direct and reliable comparisons between different architectures.
This allows us to present an up-to-date overview of the current progress in CGA models, in light of recent advancements in language modeling.
Our framework also introduces a novel metric that captures a model’s ability to revise its forecast as the conversation progresses.

\end{abstract}

\section{Introduction}
\label{sec:intro}
Our ability to foresee the likely trajectory of an ongoing discussion helps us adapt our conversational behavior. 
If we feel our interlocutor is losing interest, we can switch to more engaging topics;
if our arguments seem ineffective, we can attempt new ones;
or if we sense the tension is rising, we can be more cautious in our wording or use de-escalatory strategies.
Endowing computational systems with similar foresight can improve their ability to assist human-human conversations \cite{jurgens-etal-2019-just, 10.1145/3555603, 10.1145/3555095}.

The development of models that can predict the \textit{future} trajectory of an \textit{ongoing} conversation---i.e., conversational forecasting models---has been galvanized by the introduction of the Conversation Gone Awry (CGA) datasets \cite{zhang-etal-2018-conversations,chang-danescu-niculescu-mizil-2019-trouble}.
These datasets include conversations that either derail into personal attacks or stay civil throughout, and the task is to predict whether an \textit{ongoing} conversation will derail or not, \textit{before it does so}.
To tackle this task, models need to process conversational dynamics in real time, continuously analyzing the unfolding conversation to determine whether and when tension between interlocutors is likely to escalate.

Several architectures have been proposed, including hierarchical recurrent neural networks \cite{chang-danescu-niculescu-mizil-2019-trouble}, transformer-based models \cite{kementchedjhieva-sogaard-2021-dynamic}, hierarchical transformer-based models \cite{yuan_conversation_2023}, and graph convolutional networks \cite{altarawneh-etal-2023-conversation}.
However, in contrast to many traditional NLP tasks, this task lacks standardized evaluation benchmarks, which has stifled progress.
This absence highlights the unique evaluation challenges posed by the forecasting nature of the task, which are not encountered in traditional NLP classification tasks.

To support meaningful progress on conversational forecasting, in this work, we introduce the first standardized framework for evaluating forecasting models, released as part of ConvoKit.\footnote{\url{https://convokit.cornell.edu}}
Using this framework, we conducted the first comprehensive survey of 13 CGA models, spanning thousands to billions of parameters. This survey provides an up-to-date overview of current progress in conversation forecasting, contextualized within recent advances in language modeling.
It also reveals a new \textbf{state-of-the-art} (SOTA) \textbf{performance} for the CGA task.
Finally, our work introduces \textbf{a new metric} to address a critical gap in previous evaluations of conversational forecasting systems. 
Incorporating this metric into our evaluation framework offers a new perspective into the performance of forecasting systems.

\section{Background and Related Work}
\label{sec:related}
Conversational forecasting includes tasks with the shared goal of predicting whether an ongoing conversation will lead to some specified event or outcome.
Notable examples are forecasting whether a persuasion attempt will succeed \citep{10.1145/2872427.2883081,wachsmuth-etal-2018-retrieval,yang-etal-2019-lets}, what decision will eventually come out of an ongoing team discussion \citep{niculae_conversational_2016, 10.1145/3359308, smith_leveraging_2023}, and whether an online conversation will eventually spark disagreement \citep{hessel-lee-2019-somethings} or even outright antisocial behavior \citep{zhang-etal-2018-conversations,chang_trajectories_2019,kementchedjhieva-sogaard-2021-dynamic}.

\xhdr{Applications}
Conversational forecasting is key to several emerging real-world applications that offer \emph{proactive} conversational support.
For instance, forecasting conversation derailment has been used to alert moderators so that they can intervene before discussions get out of control \cite{10.1145/3555095}. 
It also powers applications like ConvoWizard, which gives real-time feedback to participants in online discussions by showing how likely their conversation is to derail and whether their draft reply is (de)escalatory---helping them notice tension and make changes before posting \cite{10.1145/3555603}. 
Beyond online discussions, forecasting can assist therapists in anticipating the trajectory or outcome of ongoing sessions with support seekers \cite{wang_practice_2024}, and help support chatbots track goal progress \cite{hu-etal-2022-context}.

\xhdr{Formalism: conversational forecasting models}
While forecasting may seem like a classification task predicting future events, its online nature introduces unique challenges that set it apart from traditional classification tasks.
Applying standard classification models to conversational forecasting necessarily involves a series of compromising simplifications, typically ignoring the temporal dimension by taking only a fixed snapshot of the conversation \citep{curhan_thin_2007,niculae_conversational_2016,10.1145/2872427.2883081,althoff_large-scale_2016,zhang-etal-2018-conversations,zhang_characterizing_2018}.

However, in practical scenarios, we must move beyond such compromising simplifications, especially when the temporal dimension is critical to provide \emph{real-time} guidance. 
This requires explicitly integrating the temporal aspect into the task formulation:
\begin{equation}
    \indiforecast = \mathcal{P}(\mathrm{event}\,\big|\,\mathrm{conversation}(t))
    \label{eq:forecastingmodel}
\end{equation}
Where $\mathrm{conversation}(t)$ represents the state of the conversation up to timestamp $t$, i.e., all utterances exchanged in the conversation up to that timestamp.
This means the model must update its belief about the likelihood of the event happening $\indiforecast$ (\textbf{individual forecasts}) at every timestamp, integrating information from every new utterance.
Practical applications often involve converting these likelihoods into binary predictions of whether the event will occur, $\widehat{g_t} \in \{0,1\}$.
This is typically accomplished using a threshold $T$:  $\widehat{g_t}=1$ if $\indiforecast>T$, and $\widehat{g_t}=0$ otherwise.
However, models could in principle consider other factors, such as forecasts from prior timestamps or contextual factors.

As datasets (and labels) are often organized at the conversation-level, it is common practice to aggregate individual forecasts into a single \textbf{conversation-level forecast}.
If $\widehat{g_k}=1$ at any timestamp $k$ during the conversation we say that the model \textit{triggers}: its conversation-level prediction is that the event will happen.
If the model never triggers ($\widehat{g_t}=0$ for all $t$), it implicitly predicts that the event will not happen.

\section{The Conversations Gone Awry Task}
\label{sec:data}

\begin{table*}[!t]
    \centering
    \begin{tabular}{l|cccccccl}
    \toprule
        Model & Acc $\uparrow$ & P $\uparrow$ & R $\uparrow$ & F1 $\uparrow$ & FPR $\downarrow$ & Mean H $\uparrow$ & \shortstack[l]{Recovery $\uparrow$\\($\corrcount/N - \incorrcount/N$)}\tabularnewline
    \midrule
        CRAFT& $62.8$ & $59.4$ & $\mathbf{81.1}$ & $68.5$ & $55.5$ & $\mathbf{4.7}$ & $\mathbf{+4.9}$ $\mathbf{(12.0 - 7.1)}$ \tabularnewline 
        BERT-base& $65.3$ & $64.1$ & $70.1$ & $66.9$ & $39.5$ & $4.4$ & $+1.9$ $(9.7 - 7.8)$ \tabularnewline
        RoBERTa-base& $68.1$ & $67.3$ & $70.6$ & $68.8$ & $34.4$ & $4.1$ & $+0.7$ $(7.4 - 6.7)$ \tabularnewline
        SpanBERT-base& $66.4$ & $64.7$ & $72.0$ & $68.2$ & $39.3$ & $4.4$ & $+1.7$ $(9.6 - 8.0)$ \tabularnewline
        DeBERTaV3-base& $67.9$ & $66.7$ & $71.4$ & $69.0$ & $35.7$ & $4.2$ & $+1.5$ $(7.2 - 5.7)$ \tabularnewline
        \midrule
        BERT-large& $65.7$ & $66.0$ & $65.4$ & $65.5$ & $34.1$ & $4.2$ & $+0.4$ $(7.8 - 7.3)$ \tabularnewline
        RoBERTa-large& $68.6$ & $67.1$ & $73.4$ & $70.0$ & $36.1$ & $4.2$ & $+1.6$ $(7.5 - 5.9)$ \tabularnewline
        SpanBERT-large& $67.0$ & $65.8$ & $70.5$ & $68.1$ & $36.6$ & $4.2$ & $+1.3$ $(8.3 - 7.0)$ \tabularnewline
        DeBERTaV3-large& $68.9$ & $67.3$ & $73.7$ & $70.3$ & $36.0$ & $4.2$ & $+1.1$ $(7.6 - 6.5)$ \tabularnewline
        \midrule
        Gemma2 9B& $\mathbf{71.0}$ & $\mathbf{69.1}$ & $76.1$ & $72.3$ & $34.2$ & $3.9$ & $+1.8$ $(8.4 - 6.6)$ \tabularnewline
        LlaMA3.1 8B& $70.0$ & $68.8$ & $73.2$ & $70.9$ & $\mathbf{33.2}$ & $4.0$ & $+1.7$ $(7.3 - 5.6)$ \tabularnewline
        Mistral 7B& $70.7$ & $68.8$ & $76.0$ & $72.1$ & $34.6$ & $4.0$ & $+2.9$ $(8.1 - 5.2)$ \tabularnewline
        Phi4 14B& $70.5$ & $67.7$ & $78.4$ & $\mathbf{72.6}$ & $37.5$ & $4.1$ & $+2.0$ $(7.7 - 5.7)$ \tabularnewline

        \bottomrule
    
    \end{tabular}
    \caption{%
        \textbf{Forecasting derailment on \newcmv conversations.}
        The performance is measured in accuracy (Acc), precision (P), recall (R), F1, false positive rate (FPR), mean horizon (Mean H), and Forecast Recovery (Recovery) along with the correct and incorrect recovery rates.
        Results are reported as averages over five runs with different random seeds.
    }
    \label{tab:survey-cmv_large}
\end{table*}

\xhdr{Datasets}
In this task, the event to be forecasted is a personal attack.
Previous work introduces two CGA datasets: \wiki \cite{zhang-etal-2018-conversations} and \cgacmv \cite{chang-danescu-niculescu-mizil-2019-trouble}. 
In this work, we introduce an expanded version of CGA-CMV, the \newcmv. 
To distinguish this extended version from the original \cgacmv, we will refer to the original dataset as \oldcmv from this point forward.
Previously, \citet{chang-danescu-niculescu-mizil-2019-trouble} collected \oldcmv conversations from the subreddit ChangeMyView\footnote{\href{https://www.reddit.com/r/changemyview}{reddit.com/r/changemyview}} spanning the years $2015$ to $2018$.
The labels in this dataset are collected based on whether a conversation eventually ends with a comment removed by a moderator for violating Rule 2 of ChangeMyView: ``Don’t be rude or hostile to other users.''
Building on the approach by \citet{chang-danescu-niculescu-mizil-2019-trouble}, we extend the collection period to include conversations up to $2022$, tripling the size of the data to  $19,578$ conversations.\footnote{We also make a relatively minor fix to the dataset by excluding all conversations containing deleted utterances (either by the users themselves or by Reddit for reasons other than Rule 2), 
since those no longer reflect the state of the conversation as witnessed at the time by the participants.}
We follow the same pairing and data-splitting procedures as outlined in the original papers, using a train-validation-test split ratio of 60-20-20.

Further details about our three datasets are provided in Appendix~\ref{appendix:data}.
In the following discussion, we primarily focus on the \newcmv dataset, while similar experiments conducted on \wiki and \oldcmv, for direct comparison with prior work, are reported in Appendix~\ref{appendix:otherresult}. 

\xhdr{Metrics}
For evaluation, we follow prior work and compare a model's conversation-level forecasts with (also conversation-level) ground-truth labels indicating whether the conversation contains a personal attack.
This allows classical classification metrics (e.g., accuracy, precision, recall, F1) to be applied. 
In addition to these metrics, \citet{chang-danescu-niculescu-mizil-2019-trouble} introduced \textit{mean horizon} (Mean H), which measures how early a model predicts an event before it occurs.

\section{Evaluation Framework} 
\label{sec:framework}
A reliable benchmarking framework is essential, as prior work has highlighted difficulties in comparing existing models\footnote{The performance of CRAFT in \cite{janiszewski} and \cite{kementchedjhieva-sogaard-2021-dynamic} is lower than originally reported. 
Our reproduced BERT-base models also significantly outperform the results reported by \citet{kementchedjhieva-sogaard-2021-dynamic}.}
and uses incompatible evaluation setups (e.g., some assume knowledge about the eventual length of the conversations while others do not).
Our framework adopts a modular design that abstracts evaluation aspects from model-specific components, ensuring consistent evaluation across models with different architectures.
This modularity also ensures future-proofing, facilitating the easy addition of new metrics and datasets over time.

\xhdr{Comparing forecasting architectures} 

We use this framework to provide the first up-to-date survey of CGA progress by benchmarking 13 models ranging from thousands to billions of parameters:
RNN-based CRAFT \cite{chang-danescu-niculescu-mizil-2019-trouble}, BERT \cite{devlin-etal-2019-bert}, RoBERTa \cite{liu2019robertarobustlyoptimizedbert}, SpanBERT \cite{joshi-etal-2020-spanbert}, DeBERTaV3 \cite{he2023debertav}, Gemma2 \cite{gemmateam2024gemma2improvingopen}, LlaMA3.1  \cite{grattafiori2024llama3herdmodels}, Mistral 7B \cite{jiang2023mistral7b}, and Phi4 \cite{abdin2024phi4technicalreport}.

We adopt the most effective training approach identified in prior research \cite{kementchedjhieva-sogaard-2021-dynamic, chang-danescu-niculescu-mizil-2019-trouble}, which involves training models using a snapshot of the conversation before the final comment (from which the conversation-level label is extracted, depending on whether it violated Rule 2 or not).
Importantly, during testing, the models also do not see the last utterance of the conversation, since that would give away the conversation-level outcome.
Also following prior work, we use the development split of the data to select the training checkpoint and triggering threshold $T$ that maximizes accuracy.\footnote{Our ablation study in Appendix~\ref{appendix:t-tuning} highlights the critical role of threshold tuning in this task---a factor that has not been addressed in prior work.}
Further details on model implementation and training are provided in Appendices~\ref{appendix:modelimplementation} and \ref{appendix:modeltraining}, respectively.

\xhdr{Results}
Table~\ref{tab:survey-cmv_large} presents the results for \newcmv.
We observe that Gemma2, Mistral, and Phi4 are SOTA models on \newcmv, performing similarly in terms of accuracy and F1 score. 
Across all datasets, Gemma2 achieves SOTA performance on our primary metric, accuracy (results for CGA-Wiki and \oldcmv are provided in Appendix~\ref{appendix:otherresult}).
Notably, decoder-based generative Large Language Models are the only models that are able to surpass $70\%$ accuracy.

\xhdr{Limitations of current metrics}
The traditional classification metrics (e.g., accuracy, F1) are based on conversation-level forecasts.
Therefore, they fail to account for the dynamic and online nature of the task, where forecasts are made incrementally as the conversation unfolds---one forecast for each new utterance.
The general assumption is that once a model triggers (the smallest $t$ for which $\widehat{g_t} = 1$), the rest of the individual forecasts are no longer relevant because the model has made its call.
As a result, we cannot distinguish between a model that makes forecast \emph{recoveries}---initially predicting derailment but later revising its prediction to non-derailment as new utterances are observed---and one that disregards new information and remains fixed on its initial prediction.
Therefore, traditional metrics designed for classification do not align with real-world forecasting applications. 
For example, in ConvoWizard \cite{10.1145/3555603}, if models cannot recognize when a tense conversation is recovering and instead remain fixed on their initial derailment prediction, they fail to provide meaningful feedback on users’ efforts to de-escalate tense interactions.

\xhdr{New metric: forecast \recovery}
To address this evaluation gap, we introduce a new metric designed to capture a model's ability to recover after incorrectly triggering a derailment prediction.
This metric is agnostic to whether this incorrect prediction is due to a model error or to the conversation actually recovering, since in both cases it is desirable for the forecast to recover.

A \recovery happens when the model first triggers a derailment forecast (at the first timestamp $t$ for which $\widehat{g_t}=1$) and then by the end of the conversation it changes its forecast to non-derailment ($\widehat{g_{n-1}}=0$, where $n$ is the timestamp of the label-containing utterance).
We must distinguish between \textit{correct} recoveries---when the conversation does not eventually derail---and \textit{incorrect} recoveries---when the conversation actually derails.

We design a metric for \recovery such that it favors correct {\recoveries} and discourages incorrect {\recoveries}. Letting $\corrcount$ denote the number of correct {\recoveries} and $\incorrcount$ the number of incorrect {\recoveries}, a natural metric of \recovery that meets these criteria is the difference between the fraction of correct and incorrect {\recoveries} in the test data:
\[
\mathrm{Recovery} = \frac{\corrcount}{N} - \frac{\incorrcount}{N},
\]
where $N$ is the size of the test set.
To evaluate the empirical utility of this metric, we conduct an exploratory ablation study by creating naive variants of SOTA models that lack the ability to process conversational context (deemed critical by \citet{chang-danescu-niculescu-mizil-2019-trouble}). 
While the accuracy gap between full and naive variants is within $2.5\%$, the \Recovery metric reveals a larger gap ($\sim 6\%$), going from a negative value (i.e., more incorrect than correct recoveries) to a positive one. 
Further details are presented in Appendix~\ref{appendix:recovery}.

Comparing recovery scores in Table~\ref{tab:survey-cmv_large}, we note that although Gemma2 and Mistral perform similarly according to traditional measures, Mistral 7B outperforms Gemma2 9B on the \Recovery metric.
This suggests that Mistral 7B may be better suited for real-time interactive applications like ConvoWizard \cite{10.1145/3555603}.

\section{Conclusion}
\label{sec:conclusion}
In this paper, we introduce the first standardized framework for evaluating and directly comparing forecasting models.
Using this framework, we perform the first large survey comparing 13 CGA models ranging from thousands to billions of parameters.
Our framework also addresses a key limitation in prior work: metrics borrowed from classification tasks overlook a model's ability to correct earlier mistakes as the conversations unfolds. To address this, we propose a new evaluation metric that captures these forecasting dynamics and empirically show that, when used alongside existing metrics, it provides complementary insights into the performance of forecasting models that are relevant for their use in real-world applications.

\section*{Limitations}
We acknowledge certain limitations in our work.
First, our proposed Forecast \Recovery metric overlooks \recoveries that occur at a finer granularity.
However, evaluating \recoveries at a finer granularity would require utterance-level labels, which are difficult---if not impossible---to obtain due to their highly subjective nature (they would entail judging whether the conversation is on a derailing trajectory at each moment in the conversation). 
In this work, we use the observation that the only utterance-level ground truth label that we can reliably know is the one of the last utterance, which coincides with the ground truth label of the conversation, because nothing can happen between the last utterance and the event.

Second, our survey evaluates only text-based methods, excluding information such as speaker identities and conversational graph structures \cite{yuan_conversation_2023, altarawneh_conversation_2023}. However, our open-source evaluation framework is designed to be easily extendable, allowing for the integration of these elements in future work.

Lastly, although our evaluation framework is general for conversational forecasting, we apply it only to two CGA datasets (\wiki and \cgacmv) in this work. 
Nonetheless, our open-source framework provides the basis for future development of other forecasting tasks, for example, those predicting prosocial events \cite{bao_conversations_2021}. 

\section*{Ethical Considerations}

We caution that our automated forecasters, especially decoder-based generative models like Gemma2, may unintentionally reflect biases present in the training data \citep{park_reducing_2018,sap_risk_2019,wiegand_detection_2019,chang-danescu-niculescu-mizil-2019-trouble}.
Such models are therefore most appropriately used alongside human supervision, where they may serve to supplement a human user's intuition or expertise in making proactive decisions, and not as fully autonomous systems.
\cut{
This contrasts with the bias-mitigation mechanisms commonly integrated into proprietary large language models such as GPT-$4$.
Therefore, we recommend using our models alongside human moderators to enable proactive preventative action.
}

Building on the considerations outlined by \citet{kementchedjhieva-sogaard-2021-dynamic,altarawneh_conversation_2023}, we reiterate the importance of additional considerations essential for the practical deployment of our forecasting system.
These include ensuring fairness in forecasting \citep{williamson_fairness_2019}, informing users in advance and when forecasts impact them, and determining appropriate actions upon forecasted derailment.
For a comprehensive discussion on these topics, see \citet{kiritchenko_confronting_2021}, which provides an overview of similar considerations in the context of abusive language detection.

\bibliography{refs}

\begin{thebibliography}{41}
\providecommand{\natexlab}[1]{#1}

\bibitem[{Abdin et~al.(2024)Abdin, Aneja, Behl, Bubeck, Eldan, Gunasekar, Harrison, Hewett, Javaheripi, Kauffmann, Lee, Lee, Li, Liu, Mendes, Nguyen, Price, de~Rosa, Saarikivi, Salim, Shah, Wang, Ward, Wu, Yu, Zhang, and Zhang}]{abdin2024phi4technicalreport}
Marah Abdin, Jyoti Aneja, Harkirat Behl, Sébastien Bubeck, Ronen Eldan, Suriya Gunasekar, Michael Harrison, Russell~J. Hewett, Mojan Javaheripi, Piero Kauffmann, James~R. Lee, Yin~Tat Lee, Yuanzhi Li, Weishung Liu, Caio C.~T. Mendes, Anh Nguyen, Eric Price, Gustavo de~Rosa, Olli Saarikivi, Adil Salim, Shital Shah, Xin Wang, Rachel Ward, Yue Wu, Dingli Yu, Cyril Zhang, and Yi~Zhang. 2024.
\newblock \href {https://arxiv.org/abs/2412.08905} {Phi-4 technical report}.
\newblock \emph{Preprint}, arXiv:2412.08905.

\bibitem[{Altarawneh et~al.(2023{\natexlab{a}})Altarawneh, Agrawal, Jenkin, and Papagelis}]{altarawneh-etal-2023-conversation}
Enas Altarawneh, Ameeta Agrawal, Michael Jenkin, and Manos Papagelis. 2023{\natexlab{a}}.
\newblock \href {https://doi.org/10.18653/v1/2023.woah-1.16} {Conversation derailment forecasting with graph convolutional networks}.
\newblock In \emph{The 7th Workshop on Online Abuse and Harms (WOAH)}, pages 160--169, Toronto, Canada. Association for Computational Linguistics.

\bibitem[{Altarawneh et~al.(2023{\natexlab{b}})Altarawneh, Agrawal, Jenkin, and Papagelis}]{altarawneh_conversation_2023}
Enas Altarawneh, Ameeta Agrawal, Michael Jenkin, and Manos Papagelis. 2023{\natexlab{b}}.
\newblock \href {https://doi.org/10.18653/v1/2023.woah-1.16} {Conversation {Derailment} {Forecasting} with {Graph} {Convolutional} {Networks}}.
\newblock In \emph{The 7th {Workshop} on {Online} {Abuse} and {Harms} ({WOAH})}, pages 160--169, Toronto, Canada. Association for Computational Linguistics.

\bibitem[{Althoff et~al.(2016)Althoff, Clark, and Leskovec}]{althoff_large-scale_2016}
Tim Althoff, Kevin Clark, and Jure Leskovec. 2016.
\newblock \href {https://doi.org/10.1162/tacl_a_00111} {Large-scale analysis of counseling conversations: An application of natural language processing to mental health}.
\newblock \emph{Transactions of the Association for Computational Linguistics}, 4:463--476.

\bibitem[{Bao et~al.(2021)Bao, Wu, Zhang, Chandrasekharan, and Jurgens}]{bao_conversations_2021}
Jiajun Bao, Junjie Wu, Yiming Zhang, Eshwar Chandrasekharan, and David Jurgens. 2021.
\newblock \href {https://doi.org/10.1145/3442381.3450122} {Conversations gone alright: Quantifying and predicting prosocial outcomes in online conversations}.
\newblock In \emph{Proceedings of the Web Conference 2021}, WWW '21, page 1134–1145, New York, NY, USA. Association for Computing Machinery.

\bibitem[{Chang and Danescu-Niculescu-Mizil(2019{\natexlab{a}})}]{chang_trajectories_2019}
Jonathan Chang and Cristian Danescu-Niculescu-Mizil. 2019{\natexlab{a}}.
\newblock \href {https://doi.org/10.1145/3308558.3313638} {Trajectories of blocked community members: Redemption, recidivism and departure}.
\newblock In \emph{The World Wide Web Conference}, WWW '19, page 184–195, New York, NY, USA. Association for Computing Machinery.

\bibitem[{Chang et~al.(2020)Chang, Chiam, Fu, Wang, Zhang, and Danescu-Niculescu-Mizil}]{chang-etal-2020-convokit}
Jonathan~P. Chang, Caleb Chiam, Liye Fu, Andrew Wang, Justine Zhang, and Cristian Danescu-Niculescu-Mizil. 2020.
\newblock \href {https://doi.org/10.18653/v1/2020.sigdial-1.8} {{C}onvo{K}it: A toolkit for the analysis of conversations}.
\newblock In \emph{Proceedings of the 21th Annual Meeting of the Special Interest Group on Discourse and Dialogue}, pages 57--60, 1st virtual meeting. Association for Computational Linguistics.

\bibitem[{Chang and Danescu-Niculescu-Mizil(2019{\natexlab{b}})}]{chang-danescu-niculescu-mizil-2019-trouble}
Jonathan~P. Chang and Cristian Danescu-Niculescu-Mizil. 2019{\natexlab{b}}.
\newblock \href {https://doi.org/10.18653/v1/D19-1481} {Trouble on the horizon: Forecasting the derailment of online conversations as they develop}.
\newblock In \emph{Proceedings of the 2019 Conference on Empirical Methods in Natural Language Processing and the 9th International Joint Conference on Natural Language Processing (EMNLP-IJCNLP)}, pages 4743--4754, Hong Kong, China. Association for Computational Linguistics.

\bibitem[{Chang et~al.(2022)Chang, Schluger, and Danescu-Niculescu-Mizil}]{10.1145/3555603}
Jonathan~P. Chang, Charlotte Schluger, and Cristian Danescu-Niculescu-Mizil. 2022.
\newblock \href {https://doi.org/10.1145/3555603} {Thread with caution: Proactively helping users assess and deescalate tension in their online discussions}.
\newblock \emph{Proc. ACM Hum.-Comput. Interact.}, 6(CSCW2).

\bibitem[{Curhan and Pentland(2007)}]{curhan_thin_2007}
Jared~R. Curhan and Alex Pentland. 2007.
\newblock \href {https://doi.org/10.1037/0021-9010.92.3.802} {Thin slices of negotiation: Predicting outcomes from conversational dynamics within the first 5 minutes.}
\newblock \emph{Journal of Applied Psychology}, 92:802--811.

\bibitem[{Devlin et~al.(2019)Devlin, Chang, Lee, and Toutanova}]{devlin-etal-2019-bert}
Jacob Devlin, Ming-Wei Chang, Kenton Lee, and Kristina Toutanova. 2019.
\newblock \href {https://doi.org/10.18653/v1/N19-1423} {{BERT}: Pre-training of deep bidirectional transformers for language understanding}.
\newblock In \emph{Proceedings of the 2019 Conference of the North {A}merican Chapter of the Association for Computational Linguistics: Human Language Technologies, Volume 1 (Long and Short Papers)}, pages 4171--4186, Minneapolis, Minnesota. Association for Computational Linguistics.

\bibitem[{Grattafiori et~al.(2024)Grattafiori, Dubey, Jauhri, Pandey, Kadian, Al-Dahle, Letman, Mathur, Schelten, Vaughan, Yang, Fan, Goyal, Hartshorn, Yang, Mitra, Sravankumar, Korenev, Hinsvark, Rao, Zhang, Rodriguez, Gregerson, Spataru, Roziere, Biron, Tang, Chern, Caucheteux, Nayak, Bi, Marra, McConnell, Keller, Touret, Wu, Wong, Ferrer, Nikolaidis, Allonsius, Song, Pintz, Livshits, Wyatt, Esiobu, Choudhary, Mahajan, Garcia-Olano, Perino, Hupkes, Lakomkin, AlBadawy, Lobanova, Dinan, Smith, Radenovic, Guzmán, Zhang, Synnaeve, Lee, Anderson, Thattai, Nail, Mialon, Pang, Cucurell, Nguyen, Korevaar, Xu, Touvron, Zarov, Ibarra, Kloumann, Misra, Evtimov, Zhang, Copet, Lee, Geffert, Vranes, Park, Mahadeokar, Shah, van~der Linde, Billock, Hong, Lee, Fu, Chi, Huang, Liu, Wang, Yu, Bitton, Spisak, Park, Rocca, Johnstun, Saxe, Jia, Alwala, Prasad, Upasani, Plawiak, Li, Heafield, Stone, El-Arini, Iyer, Malik, Chiu, Bhalla, Lakhotia, Rantala-Yeary, van~der Maaten, Chen, Tan, Jenkins, Martin, Madaan, Malo, Blecher,
  Landzaat, de~Oliveira, Muzzi, Pasupuleti, Singh, Paluri, Kardas, Tsimpoukelli, Oldham, Rita, Pavlova, Kambadur, Lewis, Si, Singh, Hassan, Goyal, Torabi, Bashlykov, Bogoychev, Chatterji, Zhang, Duchenne, Çelebi, Alrassy, Zhang, Li, Vasic, Weng, Bhargava, Dubal, Krishnan, Koura, Xu, He, Dong, Srinivasan, Ganapathy, Calderer, Cabral, Stojnic, Raileanu, Maheswari, Girdhar, Patel, Sauvestre, Polidoro, Sumbaly, Taylor, Silva, Hou, Wang, Hosseini, Chennabasappa, Singh, Bell, Kim, Edunov, Nie, Narang, Raparthy, Shen, Wan, Bhosale, Zhang, Vandenhende, Batra, Whitman, Sootla, Collot, Gururangan, Borodinsky, Herman, Fowler, Sheasha, Georgiou, Scialom, Speckbacher, Mihaylov, Xiao, Karn, Goswami, Gupta, Ramanathan, Kerkez, Gonguet, Do, Vogeti, Albiero, Petrovic, Chu, Xiong, Fu, Meers, Martinet, Wang, Wang, Tan, Xia, Xie, Jia, Wang, Goldschlag, Gaur, Babaei, Wen, Song, Zhang, Li, Mao, Coudert, Yan, Chen, Papakipos, Singh, Srivastava, Jain, Kelsey, Shajnfeld, Gangidi, Victoria, Goldstand, Menon, Sharma, Boesenberg,
  Baevski, Feinstein, Kallet, Sangani, Teo, Yunus, Lupu, Alvarado, Caples, Gu, Ho, Poulton, Ryan, Ramchandani, Dong, Franco, Goyal, Saraf, Chowdhury, Gabriel, Bharambe, Eisenman, Yazdan, James, Maurer, Leonhardi, Huang, Loyd, Paola, Paranjape, Liu, Wu, Ni, Hancock, Wasti, Spence, Stojkovic, Gamido, Montalvo, Parker, Burton, Mejia, Liu, Wang, Kim, Zhou, Hu, Chu, Cai, Tindal, Feichtenhofer, Gao, Civin, Beaty, Kreymer, Li, Adkins, Xu, Testuggine, David, Parikh, Liskovich, Foss, Wang, Le, Holland, Dowling, Jamil, Montgomery, Presani, Hahn, Wood, Le, Brinkman, Arcaute, Dunbar, Smothers, Sun, Kreuk, Tian, Kokkinos, Ozgenel, Caggioni, Kanayet, Seide, Florez, Schwarz, Badeer, Swee, Halpern, Herman, Sizov, Guangyi, Zhang, Lakshminarayanan, Inan, Shojanazeri, Zou, Wang, Zha, Habeeb, Rudolph, Suk, Aspegren, Goldman, Zhan, Damlaj, Molybog, Tufanov, Leontiadis, Veliche, Gat, Weissman, Geboski, Kohli, Lam, Asher, Gaya, Marcus, Tang, Chan, Zhen, Reizenstein, Teboul, Zhong, Jin, Yang, Cummings, Carvill, Shepard, McPhie,
  Torres, Ginsburg, Wang, Wu, U, Saxena, Khandelwal, Zand, Matosich, Veeraraghavan, Michelena, Li, Jagadeesh, Huang, Chawla, Huang, Chen, Garg, A, Silva, Bell, Zhang, Guo, Yu, Moshkovich, Wehrstedt, Khabsa, Avalani, Bhatt, Mankus, Hasson, Lennie, Reso, Groshev, Naumov, Lathi, Keneally, Liu, Seltzer, Valko, Restrepo, Patel, Vyatskov, Samvelyan, Clark, Macey, Wang, Hermoso, Metanat, Rastegari, Bansal, Santhanam, Parks, White, Bawa, Singhal, Egebo, Usunier, Mehta, Laptev, Dong, Cheng, Chernoguz, Hart, Salpekar, Kalinli, Kent, Parekh, Saab, Balaji, Rittner, Bontrager, Roux, Dollar, Zvyagina, Ratanchandani, Yuvraj, Liang, Alao, Rodriguez, Ayub, Murthy, Nayani, Mitra, Parthasarathy, Li, Hogan, Battey, Wang, Howes, Rinott, Mehta, Siby, Bondu, Datta, Chugh, Hunt, Dhillon, Sidorov, Pan, Mahajan, Verma, Yamamoto, Ramaswamy, Lindsay, Lindsay, Feng, Lin, Zha, Patil, Shankar, Zhang, Zhang, Wang, Agarwal, Sajuyigbe, Chintala, Max, Chen, Kehoe, Satterfield, Govindaprasad, Gupta, Deng, Cho, Virk, Subramanian, Choudhury,
  Goldman, Remez, Glaser, Best, Koehler, Robinson, Li, Zhang, Matthews, Chou, Shaked, Vontimitta, Ajayi, Montanez, Mohan, Kumar, Mangla, Ionescu, Poenaru, Mihailescu, Ivanov, Li, Wang, Jiang, Bouaziz, Constable, Tang, Wu, Wang, Wu, Gao, Kleinman, Chen, Hu, Jia, Qi, Li, Zhang, Zhang, Adi, Nam, Yu, Wang, Zhao, Hao, Qian, Li, He, Rait, DeVito, Rosnbrick, Wen, Yang, Zhao, and Ma}]{grattafiori2024llama3herdmodels}
Aaron Grattafiori, Abhimanyu Dubey, Abhinav Jauhri, Abhinav Pandey, Abhishek Kadian, Ahmad Al-Dahle, Aiesha Letman, Akhil Mathur, Alan Schelten, Alex Vaughan, Amy Yang, Angela Fan, Anirudh Goyal, Anthony Hartshorn, Aobo Yang, Archi Mitra, Archie Sravankumar, Artem Korenev, Arthur Hinsvark, Arun Rao, Aston Zhang, Aurelien Rodriguez, Austen Gregerson, Ava Spataru, Baptiste Roziere, Bethany Biron, Binh Tang, Bobbie Chern, Charlotte Caucheteux, Chaya Nayak, Chloe Bi, Chris Marra, Chris McConnell, Christian Keller, Christophe Touret, Chunyang Wu, Corinne Wong, Cristian~Canton Ferrer, Cyrus Nikolaidis, Damien Allonsius, Daniel Song, Danielle Pintz, Danny Livshits, Danny Wyatt, David Esiobu, Dhruv Choudhary, Dhruv Mahajan, Diego Garcia-Olano, Diego Perino, Dieuwke Hupkes, Egor Lakomkin, Ehab AlBadawy, Elina Lobanova, Emily Dinan, Eric~Michael Smith, Filip Radenovic, Francisco Guzmán, Frank Zhang, Gabriel Synnaeve, Gabrielle Lee, Georgia~Lewis Anderson, Govind Thattai, Graeme Nail, Gregoire Mialon, Guan Pang,
  Guillem Cucurell, Hailey Nguyen, Hannah Korevaar, Hu~Xu, Hugo Touvron, Iliyan Zarov, Imanol~Arrieta Ibarra, Isabel Kloumann, Ishan Misra, Ivan Evtimov, Jack Zhang, Jade Copet, Jaewon Lee, Jan Geffert, Jana Vranes, Jason Park, Jay Mahadeokar, Jeet Shah, Jelmer van~der Linde, Jennifer Billock, Jenny Hong, Jenya Lee, Jeremy Fu, Jianfeng Chi, Jianyu Huang, Jiawen Liu, Jie Wang, Jiecao Yu, Joanna Bitton, Joe Spisak, Jongsoo Park, Joseph Rocca, Joshua Johnstun, Joshua Saxe, Junteng Jia, Kalyan~Vasuden Alwala, Karthik Prasad, Kartikeya Upasani, Kate Plawiak, Ke~Li, Kenneth Heafield, Kevin Stone, Khalid El-Arini, Krithika Iyer, Kshitiz Malik, Kuenley Chiu, Kunal Bhalla, Kushal Lakhotia, Lauren Rantala-Yeary, Laurens van~der Maaten, Lawrence Chen, Liang Tan, Liz Jenkins, Louis Martin, Lovish Madaan, Lubo Malo, Lukas Blecher, Lukas Landzaat, Luke de~Oliveira, Madeline Muzzi, Mahesh Pasupuleti, Mannat Singh, Manohar Paluri, Marcin Kardas, Maria Tsimpoukelli, Mathew Oldham, Mathieu Rita, Maya Pavlova, Melanie Kambadur,
  Mike Lewis, Min Si, Mitesh~Kumar Singh, Mona Hassan, Naman Goyal, Narjes Torabi, Nikolay Bashlykov, Nikolay Bogoychev, Niladri Chatterji, Ning Zhang, Olivier Duchenne, Onur Çelebi, Patrick Alrassy, Pengchuan Zhang, Pengwei Li, Petar Vasic, Peter Weng, Prajjwal Bhargava, Pratik Dubal, Praveen Krishnan, Punit~Singh Koura, Puxin Xu, Qing He, Qingxiao Dong, Ragavan Srinivasan, Raj Ganapathy, Ramon Calderer, Ricardo~Silveira Cabral, Robert Stojnic, Roberta Raileanu, Rohan Maheswari, Rohit Girdhar, Rohit Patel, Romain Sauvestre, Ronnie Polidoro, Roshan Sumbaly, Ross Taylor, Ruan Silva, Rui Hou, Rui Wang, Saghar Hosseini, Sahana Chennabasappa, Sanjay Singh, Sean Bell, Seohyun~Sonia Kim, Sergey Edunov, Shaoliang Nie, Sharan Narang, Sharath Raparthy, Sheng Shen, Shengye Wan, Shruti Bhosale, Shun Zhang, Simon Vandenhende, Soumya Batra, Spencer Whitman, Sten Sootla, Stephane Collot, Suchin Gururangan, Sydney Borodinsky, Tamar Herman, Tara Fowler, Tarek Sheasha, Thomas Georgiou, Thomas Scialom, Tobias Speckbacher,
  Todor Mihaylov, Tong Xiao, Ujjwal Karn, Vedanuj Goswami, Vibhor Gupta, Vignesh Ramanathan, Viktor Kerkez, Vincent Gonguet, Virginie Do, Vish Vogeti, Vítor Albiero, Vladan Petrovic, Weiwei Chu, Wenhan Xiong, Wenyin Fu, Whitney Meers, Xavier Martinet, Xiaodong Wang, Xiaofang Wang, Xiaoqing~Ellen Tan, Xide Xia, Xinfeng Xie, Xuchao Jia, Xuewei Wang, Yaelle Goldschlag, Yashesh Gaur, Yasmine Babaei, Yi~Wen, Yiwen Song, Yuchen Zhang, Yue Li, Yuning Mao, Zacharie~Delpierre Coudert, Zheng Yan, Zhengxing Chen, Zoe Papakipos, Aaditya Singh, Aayushi Srivastava, Abha Jain, Adam Kelsey, Adam Shajnfeld, Adithya Gangidi, Adolfo Victoria, Ahuva Goldstand, Ajay Menon, Ajay Sharma, Alex Boesenberg, Alexei Baevski, Allie Feinstein, Amanda Kallet, Amit Sangani, Amos Teo, Anam Yunus, Andrei Lupu, Andres Alvarado, Andrew Caples, Andrew Gu, Andrew Ho, Andrew Poulton, Andrew Ryan, Ankit Ramchandani, Annie Dong, Annie Franco, Anuj Goyal, Aparajita Saraf, Arkabandhu Chowdhury, Ashley Gabriel, Ashwin Bharambe, Assaf Eisenman, Azadeh
  Yazdan, Beau James, Ben Maurer, Benjamin Leonhardi, Bernie Huang, Beth Loyd, Beto~De Paola, Bhargavi Paranjape, Bing Liu, Bo~Wu, Boyu Ni, Braden Hancock, Bram Wasti, Brandon Spence, Brani Stojkovic, Brian Gamido, Britt Montalvo, Carl Parker, Carly Burton, Catalina Mejia, Ce~Liu, Changhan Wang, Changkyu Kim, Chao Zhou, Chester Hu, Ching-Hsiang Chu, Chris Cai, Chris Tindal, Christoph Feichtenhofer, Cynthia Gao, Damon Civin, Dana Beaty, Daniel Kreymer, Daniel Li, David Adkins, David Xu, Davide Testuggine, Delia David, Devi Parikh, Diana Liskovich, Didem Foss, Dingkang Wang, Duc Le, Dustin Holland, Edward Dowling, Eissa Jamil, Elaine Montgomery, Eleonora Presani, Emily Hahn, Emily Wood, Eric-Tuan Le, Erik Brinkman, Esteban Arcaute, Evan Dunbar, Evan Smothers, Fei Sun, Felix Kreuk, Feng Tian, Filippos Kokkinos, Firat Ozgenel, Francesco Caggioni, Frank Kanayet, Frank Seide, Gabriela~Medina Florez, Gabriella Schwarz, Gada Badeer, Georgia Swee, Gil Halpern, Grant Herman, Grigory Sizov, Guangyi, Zhang, Guna
  Lakshminarayanan, Hakan Inan, Hamid Shojanazeri, Han Zou, Hannah Wang, Hanwen Zha, Haroun Habeeb, Harrison Rudolph, Helen Suk, Henry Aspegren, Hunter Goldman, Hongyuan Zhan, Ibrahim Damlaj, Igor Molybog, Igor Tufanov, Ilias Leontiadis, Irina-Elena Veliche, Itai Gat, Jake Weissman, James Geboski, James Kohli, Janice Lam, Japhet Asher, Jean-Baptiste Gaya, Jeff Marcus, Jeff Tang, Jennifer Chan, Jenny Zhen, Jeremy Reizenstein, Jeremy Teboul, Jessica Zhong, Jian Jin, Jingyi Yang, Joe Cummings, Jon Carvill, Jon Shepard, Jonathan McPhie, Jonathan Torres, Josh Ginsburg, Junjie Wang, Kai Wu, Kam~Hou U, Karan Saxena, Kartikay Khandelwal, Katayoun Zand, Kathy Matosich, Kaushik Veeraraghavan, Kelly Michelena, Keqian Li, Kiran Jagadeesh, Kun Huang, Kunal Chawla, Kyle Huang, Lailin Chen, Lakshya Garg, Lavender A, Leandro Silva, Lee Bell, Lei Zhang, Liangpeng Guo, Licheng Yu, Liron Moshkovich, Luca Wehrstedt, Madian Khabsa, Manav Avalani, Manish Bhatt, Martynas Mankus, Matan Hasson, Matthew Lennie, Matthias Reso, Maxim
  Groshev, Maxim Naumov, Maya Lathi, Meghan Keneally, Miao Liu, Michael~L. Seltzer, Michal Valko, Michelle Restrepo, Mihir Patel, Mik Vyatskov, Mikayel Samvelyan, Mike Clark, Mike Macey, Mike Wang, Miquel~Jubert Hermoso, Mo~Metanat, Mohammad Rastegari, Munish Bansal, Nandhini Santhanam, Natascha Parks, Natasha White, Navyata Bawa, Nayan Singhal, Nick Egebo, Nicolas Usunier, Nikhil Mehta, Nikolay~Pavlovich Laptev, Ning Dong, Norman Cheng, Oleg Chernoguz, Olivia Hart, Omkar Salpekar, Ozlem Kalinli, Parkin Kent, Parth Parekh, Paul Saab, Pavan Balaji, Pedro Rittner, Philip Bontrager, Pierre Roux, Piotr Dollar, Polina Zvyagina, Prashant Ratanchandani, Pritish Yuvraj, Qian Liang, Rachad Alao, Rachel Rodriguez, Rafi Ayub, Raghotham Murthy, Raghu Nayani, Rahul Mitra, Rangaprabhu Parthasarathy, Raymond Li, Rebekkah Hogan, Robin Battey, Rocky Wang, Russ Howes, Ruty Rinott, Sachin Mehta, Sachin Siby, Sai~Jayesh Bondu, Samyak Datta, Sara Chugh, Sara Hunt, Sargun Dhillon, Sasha Sidorov, Satadru Pan, Saurabh Mahajan,
  Saurabh Verma, Seiji Yamamoto, Sharadh Ramaswamy, Shaun Lindsay, Shaun Lindsay, Sheng Feng, Shenghao Lin, Shengxin~Cindy Zha, Shishir Patil, Shiva Shankar, Shuqiang Zhang, Shuqiang Zhang, Sinong Wang, Sneha Agarwal, Soji Sajuyigbe, Soumith Chintala, Stephanie Max, Stephen Chen, Steve Kehoe, Steve Satterfield, Sudarshan Govindaprasad, Sumit Gupta, Summer Deng, Sungmin Cho, Sunny Virk, Suraj Subramanian, Sy~Choudhury, Sydney Goldman, Tal Remez, Tamar Glaser, Tamara Best, Thilo Koehler, Thomas Robinson, Tianhe Li, Tianjun Zhang, Tim Matthews, Timothy Chou, Tzook Shaked, Varun Vontimitta, Victoria Ajayi, Victoria Montanez, Vijai Mohan, Vinay~Satish Kumar, Vishal Mangla, Vlad Ionescu, Vlad Poenaru, Vlad~Tiberiu Mihailescu, Vladimir Ivanov, Wei Li, Wenchen Wang, Wenwen Jiang, Wes Bouaziz, Will Constable, Xiaocheng Tang, Xiaojian Wu, Xiaolan Wang, Xilun Wu, Xinbo Gao, Yaniv Kleinman, Yanjun Chen, Ye~Hu, Ye~Jia, Ye~Qi, Yenda Li, Yilin Zhang, Ying Zhang, Yossi Adi, Youngjin Nam, Yu, Wang, Yu~Zhao, Yuchen Hao, Yundi
  Qian, Yunlu Li, Yuzi He, Zach Rait, Zachary DeVito, Zef Rosnbrick, Zhaoduo Wen, Zhenyu Yang, Zhiwei Zhao, and Zhiyu Ma. 2024.
\newblock \href {https://arxiv.org/abs/2407.21783} {The llama 3 herd of models}.
\newblock \emph{Preprint}, arXiv:2407.21783.

\bibitem[{He et~al.(2023)He, Gao, and Chen}]{he2023debertav}
Pengcheng He, Jianfeng Gao, and Weizhu Chen. 2023.
\newblock \href {https://openreview.net/forum?id=sE7-XhLxHA} {De{BERT}av3: Improving de{BERT}a using {ELECTRA}-style pre-training with gradient-disentangled embedding sharing}.
\newblock In \emph{The Eleventh International Conference on Learning Representations}.

\bibitem[{Hessel and Lee(2019)}]{hessel-lee-2019-somethings}
Jack Hessel and Lillian Lee. 2019.
\newblock \href {https://doi.org/10.18653/v1/N19-1166} {Something{'}s brewing! early prediction of controversy-causing posts from discussion features}.
\newblock In \emph{Proceedings of the 2019 Conference of the North {A}merican Chapter of the Association for Computational Linguistics: Human Language Technologies, Volume 1 (Long and Short Papers)}, pages 1648--1659, Minneapolis, Minnesota. Association for Computational Linguistics.

\bibitem[{Hu et~al.(2022)Hu, Lee, Xie, Yu, Smith, and Ostendorf}]{hu-etal-2022-context}
Yushi Hu, Chia-Hsuan Lee, Tianbao Xie, Tao Yu, Noah~A. Smith, and Mari Ostendorf. 2022.
\newblock \href {https://doi.org/10.18653/v1/2022.findings-emnlp.193} {In-context learning for few-shot dialogue state tracking}.
\newblock In \emph{Findings of the Association for Computational Linguistics: EMNLP 2022}, pages 2627--2643, Abu Dhabi, United Arab Emirates. Association for Computational Linguistics.

\bibitem[{Hua et~al.(2018)Hua, Danescu-Niculescu-Mizil, Taraborelli, Thain, Sorensen, and Dixon}]{hua-etal-2018-wikiconv}
Yiqing Hua, Cristian Danescu-Niculescu-Mizil, Dario Taraborelli, Nithum Thain, Jeffery Sorensen, and Lucas Dixon. 2018.
\newblock \href {https://doi.org/10.18653/v1/D18-1305} {{W}iki{C}onv: A corpus of the complete conversational history of a large online collaborative community}.
\newblock In \emph{Proceedings of the 2018 Conference on Empirical Methods in Natural Language Processing}, pages 2818--2823, Brussels, Belgium. Association for Computational Linguistics.

\bibitem[{Janiszewski et~al.(2021)Janiszewski, Lango, and Stefanowski}]{janiszewski}
Piotr Janiszewski, Mateusz Lango, and Jerzy Stefanowski. 2021.
\newblock Time aspect in making an actionable prediction of a conversation breakdown.
\newblock In \emph{Machine Learning and Knowledge Discovery in Databases. Applied Data Science Track}, pages 351--364, Cham. Springer International Publishing.

\bibitem[{Jiang et~al.(2023)Jiang, Sablayrolles, Mensch, Bamford, Chaplot, de~las Casas, Bressand, Lengyel, Lample, Saulnier, Lavaud, Lachaux, Stock, Scao, Lavril, Wang, Lacroix, and Sayed}]{jiang2023mistral7b}
Albert~Q. Jiang, Alexandre Sablayrolles, Arthur Mensch, Chris Bamford, Devendra~Singh Chaplot, Diego de~las Casas, Florian Bressand, Gianna Lengyel, Guillaume Lample, Lucile Saulnier, Lélio~Renard Lavaud, Marie-Anne Lachaux, Pierre Stock, Teven~Le Scao, Thibaut Lavril, Thomas Wang, Timothée Lacroix, and William~El Sayed. 2023.
\newblock \href {https://arxiv.org/abs/2310.06825} {Mistral 7b}.
\newblock \emph{Preprint}, arXiv:2310.06825.

\bibitem[{Joshi et~al.(2020)Joshi, Chen, Liu, Weld, Zettlemoyer, and Levy}]{joshi-etal-2020-spanbert}
Mandar Joshi, Danqi Chen, Yinhan Liu, Daniel~S. Weld, Luke Zettlemoyer, and Omer Levy. 2020.
\newblock \href {https://doi.org/10.1162/tacl_a_00300} {{S}pan{BERT}: Improving pre-training by representing and predicting spans}.
\newblock \emph{Transactions of the Association for Computational Linguistics}, 8:64--77.

\bibitem[{Jurgens et~al.(2019)Jurgens, Hemphill, and Chandrasekharan}]{jurgens-etal-2019-just}
David Jurgens, Libby Hemphill, and Eshwar Chandrasekharan. 2019.
\newblock \href {https://doi.org/10.18653/v1/P19-1357} {A just and comprehensive strategy for using {NLP} to address online abuse}.
\newblock In \emph{Proceedings of the 57th Annual Meeting of the Association for Computational Linguistics}, pages 3658--3666, Florence, Italy. Association for Computational Linguistics.

\bibitem[{Kementchedjhieva and S{\o}gaard(2021)}]{kementchedjhieva-sogaard-2021-dynamic}
Yova Kementchedjhieva and Anders S{\o}gaard. 2021.
\newblock \href {https://doi.org/10.18653/v1/2021.emnlp-main.624} {Dynamic forecasting of conversation derailment}.
\newblock In \emph{Proceedings of the 2021 Conference on Empirical Methods in Natural Language Processing}, pages 7915--7919, Online and Punta Cana, Dominican Republic. Association for Computational Linguistics.

\bibitem[{Kiritchenko et~al.(2021)Kiritchenko, Nejadgholi, and Fraser}]{kiritchenko_confronting_2021}
Svetlana Kiritchenko, Isar Nejadgholi, and Kathleen~C. Fraser. 2021.
\newblock \href {https://doi.org/10.1613/jair.1.12590} {Confronting {Abusive} {Language} {Online}: {A} {Survey} from the {Ethical} and {Human} {Rights} {Perspective}}.
\newblock \emph{Journal of Artificial Intelligence Research}, 71:431--478.

\bibitem[{Liu et~al.(2019)Liu, Ott, Goyal, Du, Joshi, Chen, Levy, Lewis, Zettlemoyer, and Stoyanov}]{liu2019robertarobustlyoptimizedbert}
Yinhan Liu, Myle Ott, Naman Goyal, Jingfei Du, Mandar Joshi, Danqi Chen, Omer Levy, Mike Lewis, Luke Zettlemoyer, and Veselin Stoyanov. 2019.
\newblock \href {https://arxiv.org/abs/1907.11692} {Roberta: A robustly optimized bert pretraining approach}.
\newblock \emph{Preprint}, arXiv:1907.11692.

\bibitem[{Mayfield and Black(2019)}]{10.1145/3359308}
Elijah Mayfield and Alan~W. Black. 2019.
\newblock \href {https://doi.org/10.1145/3359308} {Analyzing wikipedia deletion debates with a group decision-making forecast model}.
\newblock \emph{Proc. ACM Hum.-Comput. Interact.}, 3(CSCW).

\bibitem[{Niculae and Danescu-Niculescu-Mizil(2016)}]{niculae_conversational_2016}
Vlad Niculae and Cristian Danescu-Niculescu-Mizil. 2016.
\newblock \href {https://doi.org/10.18653/v1/N16-1070} {Conversational markers of constructive discussions}.
\newblock In \emph{Proceedings of the 2016 Conference of the North {A}merican Chapter of the Association for Computational Linguistics: Human Language Technologies}, pages 568--578, San Diego, California. Association for Computational Linguistics.

\bibitem[{Park et~al.(2018)Park, Shin, and Fung}]{park_reducing_2018}
Ji~Ho Park, Jamin Shin, and Pascale Fung. 2018.
\newblock \href {https://doi.org/10.18653/v1/D18-1302} {Reducing {Gender} {Bias} in {Abusive} {Language} {Detection}}.
\newblock In \emph{Proceedings of the 2018 {Conference} on {Empirical} {Methods} in {Natural} {Language} {Processing}}, pages 2799--2804, Brussels, Belgium. Association for Computational Linguistics.

\bibitem[{Paszke et~al.(2019)Paszke, Gross, Massa, Lerer, Bradbury, Chanan, Killeen, Lin, Gimelshein, Antiga, Desmaison, K\"{o}pf, Yang, DeVito, Raison, Tejani, Chilamkurthy, Steiner, Fang, Bai, and Chintala}]{10.5555/3454287.3455008}
Adam Paszke, Sam Gross, Francisco Massa, Adam Lerer, James Bradbury, Gregory Chanan, Trevor Killeen, Zeming Lin, Natalia Gimelshein, Luca Antiga, Alban Desmaison, Andreas K\"{o}pf, Edward Yang, Zach DeVito, Martin Raison, Alykhan Tejani, Sasank Chilamkurthy, Benoit Steiner, Lu~Fang, Junjie Bai, and Soumith Chintala. 2019.
\newblock Pytorch: an imperative style, high-performance deep learning library.
\newblock In \emph{Proceedings of the 33rd International Conference on Neural Information Processing Systems}, Red Hook, NY, USA. Curran Associates Inc.

\bibitem[{Riviere et~al.(2024)Riviere, Pathak, Sessa, Hardin, Bhupatiraju, Hussenot, Mesnard, Shahriari, Ramé, Ferret, Liu, Tafti, Friesen, Casbon, Ramos, Kumar, Lan, Jerome, Tsitsulin, Vieillard, Stanczyk, Girgin, Momchev, Hoffman, Thakoor, Grill, Neyshabur, Bachem, Walton, Severyn, Parrish, Ahmad, Hutchison, Abdagic, Carl, Shen, Brock, Coenen, Laforge, Paterson, Bastian, Piot, Wu, Royal, Chen, Kumar, Perry, Welty, Choquette-Choo, Sinopalnikov, Weinberger, Vijaykumar, Rogozińska, Herbison, Bandy, Wang, Noland, Moreira, Senter, Eltyshev, Visin, Rasskin, Wei, Cameron, Martins, Hashemi, Klimczak-Plucińska, Batra, Dhand, Nardini, Mein, Zhou, Svensson, Stanway, Chan, Zhou, Carrasqueira, Iljazi, Becker, Fernandez, van Amersfoort, Gordon, Lipschultz, Newlan, yeong Ji, Mohamed, Badola, Black, Millican, McDonell, Nguyen, Sodhia, Greene, Sjoesund, Usui, Sifre, Heuermann, Lago, McNealus, Soares, Kilpatrick, Dixon, Martins, Reid, Singh, Iverson, Görner, Velloso, Wirth, Davidow, Miller, Rahtz, Watson, Risdal, Kazemi,
  Moynihan, Zhang, Kahng, Park, Rahman, Khatwani, Dao, Bardoliwalla, Devanathan, Dumai, Chauhan, Wahltinez, Botarda, Barnes, Barham, Michel, Jin, Georgiev, Culliton, Kuppala, Comanescu, Merhej, Jana, Rokni, Agarwal, Mullins, Saadat, Carthy, Cogan, Perrin, Arnold, Krause, Dai, Garg, Sheth, Ronstrom, Chan, Jordan, Yu, Eccles, Hennigan, Kocisky, Doshi, Jain, Yadav, Meshram, Dharmadhikari, Barkley, Wei, Ye, Han, Kwon, Xu, Shen, Gong, Wei, Cotruta, Kirk, Rao, Giang, Peran, Warkentin, Collins, Barral, Ghahramani, Hadsell, Sculley, Banks, Dragan, Petrov, Vinyals, Dean, Hassabis, Kavukcuoglu, Farabet, Buchatskaya, Borgeaud, Fiedel, Joulin, Kenealy, Dadashi, and Andreev}]{gemmateam2024gemma2improvingopen}
Morgane Riviere, Shreya Pathak, Pier~Giuseppe Sessa, Cassidy Hardin, Surya Bhupatiraju, Léonard Hussenot, Thomas Mesnard, Bobak Shahriari, Alexandre Ramé, Johan Ferret, Peter Liu, Pouya Tafti, Abe Friesen, Michelle Casbon, Sabela Ramos, Ravin Kumar, Charline~Le Lan, Sammy Jerome, Anton Tsitsulin, Nino Vieillard, Piotr Stanczyk, Sertan Girgin, Nikola Momchev, Matt Hoffman, Shantanu Thakoor, Jean-Bastien Grill, Behnam Neyshabur, Olivier Bachem, Alanna Walton, Aliaksei Severyn, Alicia Parrish, Aliya Ahmad, Allen Hutchison, Alvin Abdagic, Amanda Carl, Amy Shen, Andy Brock, Andy Coenen, Anthony Laforge, Antonia Paterson, Ben Bastian, Bilal Piot, Bo~Wu, Brandon Royal, Charlie Chen, Chintu Kumar, Chris Perry, Chris Welty, Christopher~A. Choquette-Choo, Danila Sinopalnikov, David Weinberger, Dimple Vijaykumar, Dominika Rogozińska, Dustin Herbison, Elisa Bandy, Emma Wang, Eric Noland, Erica Moreira, Evan Senter, Evgenii Eltyshev, Francesco Visin, Gabriel Rasskin, Gary Wei, Glenn Cameron, Gus Martins, Hadi Hashemi,
  Hanna Klimczak-Plucińska, Harleen Batra, Harsh Dhand, Ivan Nardini, Jacinda Mein, Jack Zhou, James Svensson, Jeff Stanway, Jetha Chan, Jin~Peng Zhou, Joana Carrasqueira, Joana Iljazi, Jocelyn Becker, Joe Fernandez, Joost van Amersfoort, Josh Gordon, Josh Lipschultz, Josh Newlan, Ju~yeong Ji, Kareem Mohamed, Kartikeya Badola, Kat Black, Katie Millican, Keelin McDonell, Kelvin Nguyen, Kiranbir Sodhia, Kish Greene, Lars~Lowe Sjoesund, Lauren Usui, Laurent Sifre, Lena Heuermann, Leticia Lago, Lilly McNealus, Livio~Baldini Soares, Logan Kilpatrick, Lucas Dixon, Luciano Martins, Machel Reid, Manvinder Singh, Mark Iverson, Martin Görner, Mat Velloso, Mateo Wirth, Matt Davidow, Matt Miller, Matthew Rahtz, Matthew Watson, Meg Risdal, Mehran Kazemi, Michael Moynihan, Ming Zhang, Minsuk Kahng, Minwoo Park, Mofi Rahman, Mohit Khatwani, Natalie Dao, Nenshad Bardoliwalla, Nesh Devanathan, Neta Dumai, Nilay Chauhan, Oscar Wahltinez, Pankil Botarda, Parker Barnes, Paul Barham, Paul Michel, Pengchong Jin, Petko Georgiev,
  Phil Culliton, Pradeep Kuppala, Ramona Comanescu, Ramona Merhej, Reena Jana, Reza~Ardeshir Rokni, Rishabh Agarwal, Ryan Mullins, Samaneh Saadat, Sara~Mc Carthy, Sarah Cogan, Sarah Perrin, Sébastien M.~R. Arnold, Sebastian Krause, Shengyang Dai, Shruti Garg, Shruti Sheth, Sue Ronstrom, Susan Chan, Timothy Jordan, Ting Yu, Tom Eccles, Tom Hennigan, Tomas Kocisky, Tulsee Doshi, Vihan Jain, Vikas Yadav, Vilobh Meshram, Vishal Dharmadhikari, Warren Barkley, Wei Wei, Wenming Ye, Woohyun Han, Woosuk Kwon, Xiang Xu, Zhe Shen, Zhitao Gong, Zichuan Wei, Victor Cotruta, Phoebe Kirk, Anand Rao, Minh Giang, Ludovic Peran, Tris Warkentin, Eli Collins, Joelle Barral, Zoubin Ghahramani, Raia Hadsell, D.~Sculley, Jeanine Banks, Anca Dragan, Slav Petrov, Oriol Vinyals, Jeff Dean, Demis Hassabis, Koray Kavukcuoglu, Clement Farabet, Elena Buchatskaya, Sebastian Borgeaud, Noah Fiedel, Armand Joulin, Kathleen Kenealy, Robert Dadashi, and Alek Andreev. 2024.
\newblock \href {https://arxiv.org/abs/2408.00118} {Gemma 2: Improving open language models at a practical size}.
\newblock \emph{Preprint}, arXiv:2408.00118.

\bibitem[{Sap et~al.(2019)Sap, Card, Gabriel, Choi, and Smith}]{sap_risk_2019}
Maarten Sap, Dallas Card, Saadia Gabriel, Yejin Choi, and Noah~A. Smith. 2019.
\newblock \href {https://doi.org/10.18653/v1/P19-1163} {The {Risk} of {Racial} {Bias} in {Hate} {Speech} {Detection}}.
\newblock In \emph{Proceedings of the 57th {Annual} {Meeting} of the {Association} for {Computational} {Linguistics}}, pages 1668--1678, Florence, Italy. Association for Computational Linguistics.

\bibitem[{Schluger et~al.(2022)Schluger, Chang, Danescu-Niculescu-Mizil, and Levy}]{10.1145/3555095}
Charlotte Schluger, Jonathan~P. Chang, Cristian Danescu-Niculescu-Mizil, and Karen Levy. 2022.
\newblock \href {https://doi.org/10.1145/3555095} {Proactive moderation of online discussions: Existing practices and the potential for algorithmic support}.
\newblock \emph{Proc. ACM Hum.-Comput. Interact.}, 6(CSCW2).

\bibitem[{Smith(2023)}]{smith_leveraging_2023}
Ana Smith. 2023.
\newblock \href {https://www.proquest.com/docview/2827129304/abstract/94C45D7E632D41AFPQ/1} {\emph{Leveraging Context Documents for Social Natural Language Processing}}.
\newblock phdthesis, Cornell University.
\newblock {ISBN}: 9798379712426.

\bibitem[{Tan et~al.(2016)Tan, Niculae, Danescu-Niculescu-Mizil, and Lee}]{10.1145/2872427.2883081}
Chenhao Tan, Vlad Niculae, Cristian Danescu-Niculescu-Mizil, and Lillian Lee. 2016.
\newblock \href {https://doi.org/10.1145/2872427.2883081} {Winning arguments: Interaction dynamics and persuasion strategies in good-faith online discussions}.
\newblock In \emph{Proceedings of the 25th International Conference on World Wide Web}, WWW '16, page 613–624, Republic and Canton of Geneva, CHE. International World Wide Web Conferences Steering Committee.

\bibitem[{Wachsmuth et~al.(2018)Wachsmuth, Syed, and Stein}]{wachsmuth-etal-2018-retrieval}
Henning Wachsmuth, Shahbaz Syed, and Benno Stein. 2018.
\newblock \href {https://doi.org/10.18653/v1/P18-1023} {Retrieval of the best counterargument without prior topic knowledge}.
\newblock In \emph{Proceedings of the 56th Annual Meeting of the Association for Computational Linguistics (Volume 1: Long Papers)}, pages 241--251, Melbourne, Australia. Association for Computational Linguistics.

\bibitem[{Wang et~al.(2025)Wang, Bruckman, and Yang}]{wang_practice_2024}
Tony Wang, Amy~S Bruckman, and Diyi Yang. 2025.
\newblock \href {https://doi.org/10.1145/3711089} {The {Practice} of {Online} {Peer} {Counseling} and the {Potential} for {AI}-{Powered} {Support} {Tools}}.
\newblock \emph{Proc. ACM Hum.-Comput. Interact.}, 9(CSCW).

\bibitem[{Wiegand et~al.(2019)Wiegand, Ruppenhofer, and Kleinbauer}]{wiegand_detection_2019}
Michael Wiegand, Josef Ruppenhofer, and Thomas Kleinbauer. 2019.
\newblock \href {https://doi.org/10.18653/v1/N19-1060} {Detection of {Abusive} {Language}: the {Problem} of {Biased} {Datasets}}.
\newblock In \emph{Proceedings of the 2019 {Conference} of the {North} {American} {Chapter} of the {Association} for {Computational} {Linguistics}: {Human} {Language} {Technologies}, {Volume} 1 ({Long} and {Short} {Papers})}, pages 602--608, Minneapolis, Minnesota. Association for Computational Linguistics.

\bibitem[{Williamson and Menon(2019)}]{williamson_fairness_2019}
Robert Williamson and Aditya Menon. 2019.
\newblock \href {https://proceedings.mlr.press/v97/williamson19a.html} {Fairness risk measures}.
\newblock In \emph{Proceedings of the 36th {International} {Conference} on {Machine} {Learning}}, pages 6786--6797. PMLR.
\newblock ISSN: 2640-3498.

\bibitem[{Wolf et~al.(2020)Wolf, Debut, Sanh, Chaumond, Delangue, Moi, Cistac, Rault, Louf, Funtowicz, Davison, Shleifer, von Platen, Ma, Jernite, Plu, Xu, Le~Scao, Gugger, Drame, Lhoest, and Rush}]{wolf-etal-2020-transformers}
Thomas Wolf, Lysandre Debut, Victor Sanh, Julien Chaumond, Clement Delangue, Anthony Moi, Pierric Cistac, Tim Rault, Remi Louf, Morgan Funtowicz, Joe Davison, Sam Shleifer, Patrick von Platen, Clara Ma, Yacine Jernite, Julien Plu, Canwen Xu, Teven Le~Scao, Sylvain Gugger, Mariama Drame, Quentin Lhoest, and Alexander Rush. 2020.
\newblock \href {https://doi.org/10.18653/v1/2020.emnlp-demos.6} {Transformers: State-of-the-art natural language processing}.
\newblock In \emph{Proceedings of the 2020 Conference on Empirical Methods in Natural Language Processing: System Demonstrations}, pages 38--45, Online. Association for Computational Linguistics.

\bibitem[{Yang et~al.(2019)Yang, Chen, Yang, Jurafsky, and Hovy}]{yang-etal-2019-lets}
Diyi Yang, Jiaao Chen, Zichao Yang, Dan Jurafsky, and Eduard Hovy. 2019.
\newblock \href {https://doi.org/10.18653/v1/N19-1364} {Let{'}s make your request more persuasive: Modeling persuasive strategies via semi-supervised neural nets on crowdfunding platforms}.
\newblock In \emph{Proceedings of the 2019 Conference of the North {A}merican Chapter of the Association for Computational Linguistics: Human Language Technologies, Volume 1 (Long and Short Papers)}, pages 3620--3630, Minneapolis, Minnesota. Association for Computational Linguistics.

\bibitem[{Yuan and Singh(2023)}]{yuan_conversation_2023}
Jiaqing Yuan and Munindar~P. Singh. 2023.
\newblock \href {https://doi.org/10.1609/icwsm.v17i1.22200} {Conversation modeling to predict derailment}.
\newblock \emph{Proceedings of the International AAAI Conference on Web and Social Media}, 17(1):926--935.

\bibitem[{Zhang et~al.(2018{\natexlab{a}})Zhang, Chang, Danescu-Niculescu-Mizil, Dixon, Hua, Taraborelli, and Thain}]{zhang-etal-2018-conversations}
Justine Zhang, Jonathan Chang, Cristian Danescu-Niculescu-Mizil, Lucas Dixon, Yiqing Hua, Dario Taraborelli, and Nithum Thain. 2018{\natexlab{a}}.
\newblock \href {https://doi.org/10.18653/v1/P18-1125} {Conversations gone awry: Detecting early signs of conversational failure}.
\newblock In \emph{Proceedings of the 56th Annual Meeting of the Association for Computational Linguistics (Volume 1: Long Papers)}, pages 1350--1361, Melbourne, Australia. Association for Computational Linguistics.

\bibitem[{Zhang et~al.(2018{\natexlab{b}})Zhang, Danescu-Niculescu-Mizil, Sauper, and Taylor}]{zhang_characterizing_2018}
Justine Zhang, Cristian Danescu-Niculescu-Mizil, Christina Sauper, and Sean~J. Taylor. 2018{\natexlab{b}}.
\newblock \href {https://doi.org/10.1145/3274467} {Characterizing online public discussions through patterns of participant interactions}.
\newblock In \emph{Proceedings of {CSCW}}.

\end{thebibliography}

\newpage\hbox{}\thispagestyle{empty}

\appendix

\section{Further Details on Data}
\label{appendix:data}
Following \citet{chang-danescu-niculescu-mizil-2019-trouble}, we use the same two derailment datasets for our benchmarking experiments.

\xhdr{\wiki}
\wiki \citep{chang-danescu-niculescu-mizil-2019-trouble} is an expanded version of the annotated Wikipedia dataset introduced by \citet{zhang-etal-2018-conversations}.
It consists of conversations between editors on Wikipedia Talk Pages, sourced from the WikiConv dataset \citep{hua-etal-2018-wikiconv}. 
Each conversation has been carefully annotated by crowdworkers to determine whether it contains a personal attack or remains civil throughout (derail or not).
To mitigate biases related to topic-specific information and conversational length, the dataset pairs each derailing conversation with a similarly long, non-derailing conversation from the same talk page.
This pairing ensures both topic control and class balance.
The final dataset comprises $4,188$ conversations, with a pair-preserving train-validation-test split of $60$-$20$-$20$.


\xhdr{\oldcmv}
\cgacmv \citep{chang-danescu-niculescu-mizil-2019-trouble} consists of conversations from the ChangeMyView subreddit (from $2015$--$2017$), a platform where users seek to have their views challenged on controversial topics.

While this subreddit promotes constructive dialogue, the inherently adversarial nature of debates can sometimes escalate into rude or hostile exchanges.
Unlike \wiki, which relies on post-hoc human annotations, \cgacmv leverages comment deletions by platform moderators as \cut{implicit} derailment signals.
Specifically, comments removed for violating Rule~$2$---``Don't be rude or hostile to other users''---serve as \cut{partial} annotations of derailment.

The conversations in the dataset are also paired to ensure topic control, length matching, and class balancing.
However, as \citet{chang-danescu-niculescu-mizil-2019-trouble} note, there is no explicit control to guarantee that all turns in a conversation preceding the final one remain civil.
This introduces a potential source of noise in the dataset, as earlier turns may already exhibit derailment before the moderation-triggering comment.
The final dataset consists of $6,842$ conversations, with a pair-preserving train-validation-test split of $60$-$20$-$20$.

\xhdr{\newcmv}
In this work, we extend the \cgacmv dataset by expanding the collection period to include conversations up to $2022,$ resulting in a substantially large dataset with $19,578$ conversations.
As in the \oldcmv, we ensure topic control by pairing conversation branches that contain a removed comment with those that do not, while ensuring both originate from the same top-level post. 
Additionally, we ensure length matching and class balancing to maintain a controlled comparison between derailment and non-derailment cases.
A key distinction from the \oldcmv dataset is that we exclude all conversations containing deleted utterances, whether removed by users or by Reddit's automated systems.\footnote{This ensures that deletions in our dataset strictly correspond to moderator interventions rather than user actions or platform-wide automated removals.}

We focus on \cgacmv because it represents a setting where derailment decisions are made in real-time by moderators rather than through retrospective annotation.
This aligns with our goal of evaluating derailment detection in environments where intervention must occur dynamically.
To this end, this work primarily utilizes \newcmv
(with a pair-preserving train-validation-test split of $60$-$20$-$20$), as its larger scale and more recent data offer a more comprehensive benchmark for derailment detection.
\section{Other Results}
\label{appendix:otherresult}
\begin{table*}[!ht]
    \centering
    \begin{tabular}{l|ccccccl}
    \toprule
        Model & Acc $\uparrow$ & P $\uparrow$ & R $\uparrow$ & F1 $\uparrow$ & FPR $\downarrow$ & Mean H $\uparrow$ & \shortstack[l]{Recovery $\uparrow$\\($\corrcount/N - \incorrcount/N$)}\tabularnewline
    \midrule
        CRAFT & $64.8$ & $63.4$ & $70.1$ & $66.5$ & $40.5$ & $3.5$ & $+0.4$ $(4.5 - 4.1)$\tabularnewline 
        BERT-base& $66.5$ & $66.5$ & $66.3$ & $66.4$ & $33.4$ & $3.6$ & $-1.6$ $(5.6 - 7.2)$ \tabularnewline
        RoBERTa-base& $67.6$ & $65.7$ & $73.9$ & $69.5$ & $38.6$ & $3.6$ & $+0.5$ $(3.4 - 2.8)$ \tabularnewline
        SpanBERT-base& $66.7$ & $66.1$ & $68.7$ & $67.3$ & $35.2$ & $3.3$ & $-0.7$ $(4.5 - 5.2)$ \tabularnewline
        DeBERTaV3-base& $67.5$ & $67.0$ & $69.2$ & $68.0$ & $34.3$ & $3.6$ & $+0.5$ $(2.7 - 2.3)$ \tabularnewline
        \midrule
        BERT-large& $65.7$ & $65.6$ & $67.0$ & $66.0$ & $35.6$ & $3.6$ & $+0.0$ $(5.6 - 5.6)$ \tabularnewline
        RoBERTa-large& $68.2$ & $67.8$ & $69.7$ & $68.6$ & $33.3$ & $3.6$ & $+0.3$ $(3.9 - 3.5)$ \tabularnewline
        SpanBERT-large& $67.9$ & $66.5$ & $72.6$ & $69.3$ & $36.7$ & $3.6$ & $+0.1$ $(4.9 - 4.8)$ \tabularnewline
        DeBERTaV3-large& $67.8$ & $66.9$ & $70.9$ & $68.7$ & $35.3$ & $3.7$ & $+0.8$ $(3.8 - 3.0)$ \tabularnewline
        \midrule
        Gemma2 9B& $\mathbf{69.2}$ & $67.5$ & $75.3$ & $\mathbf{70.9}$ & $36.9$ & $3.6$ & $+0.9$ $(4.1 - 3.2)$ \tabularnewline
        LlaMA3.1 8B& $68.5$ & $66.3$ & $\mathbf{75.6}$ & $70.5$ & $38.7$ & $3.6$ & $\mathbf{+1.8}$ $\mathbf{(5.5 - 3.7)}$ \tabularnewline
        Mistral 7B& $67.8$ & $65.9$ & $74.4$ & $69.8$ & $38.8$ & $\mathbf{3.8}$ & $+1.1$ $(5.1 - 4.0)$ \tabularnewline
        Phi4 14B& $68.8$ & $\mathbf{69.5}$ & $67.1$ & $68.2$ & $\mathbf{29.6}$ & $3.3$ & $+0.8$ $(3.7 - 2.9)$ \tabularnewline

    \bottomrule
    \end{tabular}
    \caption{%
        \textbf{Forecasting derailment on CGA-Wiki conversations.}
        The performance is measured in accuracy (Acc), precision (P), recall (R), F1, false positive rate (FPR), mean horizon (Mean H), and Forecast Recovery (Recovery) along with the correct and incorrect recovery rates.
        Results are reported as averages over five runs with different random seeds.
        The highest performance for each metric within each group is highlighted in \textbf{bold}.
    }
    \label{tab:survey-wikiconv}
\end{table*}

\begin{table*}[!ht]
    \centering
    \begin{tabular}{l|ccccccl}
    \toprule
        Model & Acc $\uparrow$ & P $\uparrow$ & R $\uparrow$ & F1 $\uparrow$ & FPR $\downarrow$ & Mean H $\uparrow$ & \shortstack[l]{Recovery $\uparrow$\\($\corrcount/N - \incorrcount/N$)}\tabularnewline
    \midrule
        CRAFT& $61.8$ & $59.4$ & $74.9$ & $66.2$ & $51.3$ & $\mathbf{4.7}$ & $\mathbf{+2.0}$ $\mathbf{(9.4 - 7.4)}$\tabularnewline 
        BERT-base& $63.1$ & $64.0$ & $60.8$ & $62.1$ & $34.6$ & $4.2$ & $+0.9$ $(8.1 - 7.2)$ \tabularnewline
        RoBERTa-base& $67.4$ & $65.6$ & $73.1$ & $69.1$ & $38.3$ & $4.3$ & $+2.1$ $(7.0 - 5.0)$ \tabularnewline
        SpanBERT-base& $65.7$ & $65.0$ & $68.5$ & $66.6$ & $37.0$ & $4.3$ & $+0.9$ $(7.7 - 6.8)$ \tabularnewline
        DeBERTaV3-base& $66.7$ & $65.8$ & $70.2$ & $67.8$ & $36.8$ & $4.4$ & $+1.6$ $(7.1 - 5.5)$ \tabularnewline
        \midrule
        BERT-large& $64.1$ & $64.2$ & $63.8$ & $64.0$ & $35.6$ & $4.3$ & $+0.6$ $(7.9 - 7.3)$ \tabularnewline
        RoBERTa-large& $68.3$ & $67.7$ & $70.2$ & $68.8$ & $33.7$ & $4.1$ & $+0.7$ $(5.9 - 5.2)$ \tabularnewline
        SpanBERT-large& $65.9$ & $65.5$ & $67.1$ & $66.3$ & $35.3$ & $4.1$ & $+1.2$ $(6.9 - 5.7)$ \tabularnewline
        DeBERTaV3-large& $67.4$ & $67.5$ & $67.1$ & $67.3$ & $32.4$ & $4.2$ & $+0.6$ $(5.2 - 4.6)$ \tabularnewline
        \midrule
        Gemma2 9B& $\mathbf{71.9}$ & $\mathbf{69.9}$ & $\mathbf{76.8}$ & $\mathbf{73.2}$ & $33.1$ & $4.0$ & $+1.1$ $(5.5 - 4.4)$ \tabularnewline
        LlaMA3.1 8B& $69.6$ & $68.5$ & $72.7$ & $70.5$ & $33.5$ & $4.0$ & $+1.3$ $(5.5 - 4.2)$ \tabularnewline
        Mistral 7B& $68.2$ & $68.5$ & $68.4$ & $68.0$ & $\mathbf{32.0}$ & $3.9$ & $+1.3$ $(4.8 - 3.5)$ \tabularnewline
        Phi4 14B& $70.3$ & $67.6$ & $78.4$ & $72.5$ & $37.7$ & $4.1$ & $+1.8$ $(6.8 - 5.1)$ \tabularnewline
    \bottomrule
    \end{tabular}
    \caption{%
        \textbf{Forecasting derailment on \oldcmv conversations.}
        The performance is measured in accuracy (Acc), precision (P), recall (R), F1, false positive rate (FPR), mean horizon (Mean H), and Forecast Recovery (Recovery) along with the correct and incorrect recovery rates.
        Results are reported as averages over five runs with different random seeds.
        The highest performance for each metric is highlighted in \textbf{bold}.
    }
    \label{tab:survey-cmv_legacy}
\end{table*}

Table~\ref{tab:survey-wikiconv} and Table~\ref{tab:survey-cmv_legacy} present the performance of 13 models on CGA-Wiki and \oldcmv. 

\section{Further Insights into Forecast \Recovery}
\label{appendix:recovery}
\begin{table*}[ht]
\renewcommand{\arraystretch}{1.0}
    \fontsize{10pt}{11pt}\selectfont
    \begin{tabularx}{\linewidth}{p{0.2cm}p{0.2cm}X}
    \textbf{\textcolor{red}{A}} & \textbf{\textcolor{blue}{B}} & \\
        \hline
        \textbf{\textcolor{red}{1}}&\textbf{\textcolor{red}{1}}&\textbf{\textit{(1) User1:}} Your entire premise is a non sequitur.  Intelligence is an end in itself.  It is not a benefit because it is useful.\\
        
        \textbf{\textcolor{red}{1}}&0&\textbf{\textit{(2) User2:}} What do you mean? I'm not saying all intelligence is worthless, just that the ability to memorize facts is valued very highly when in everyday life it doesn't come up very often.\\

        \textbf{\textcolor{red}{1}}&0&\textbf{\textit{(3) User3:}} Where in everyday life is it valued?\\

        \textbf{\textcolor{red}{1}}&0&\textbf{\textit{(4) User2:}} People view people who can quote poetry or who have read The Art of War as intellectuals when really then just read a book. I'm not saying there's anything wrong with that, it just isn't useful.\\
        
        &&\textbf{Conversation Label: \textcolor{teal}{Not derailed}} (conversation does not end with a personal attack)\\
    
        \hline
        0&0&\noindent\textbf{\textit{ (1) User1:}} A poodle is the same species as a wolf and they can even interbreed, but they sure are different animals. With humans, races can be equally different. [...]\\
        \textbf{\textcolor{red}{1}}&\textbf{\textcolor{red}{1}}&\noindent\textbf{\textit{ (2) User2:}} Absolute nonsense. They aren’t equally different at all.
        \\
        0&\textbf{\textcolor{red}{1}}&\noindent\textbf{\textit{ (3) User1:}} I am 6'4'' 230lbs with mostly german ancestry
        The average Filipino man is more than a foot shorter than me and half my weight.
        Yeah, I would say that we are about as different as a wolf and a poodle.\\

        



        
        
        
        &&\textbf{Conversation label: \textcolor{red}{Derailed}} (the conversation ends with a personal attack)\\
        
        \hline

    \end{tabularx}

\caption{Examples of conversations where forecast recovery may happen, illustrated using two hypothetical {\forecastingmodel}s, \textbf{\textcolor{red}{A}} and \textbf{\textcolor{blue}{B}}, where \textbf{\textcolor{blue}{B}} exhibits the desirable behaviors.}
\label{tab:recovery-example}
\end{table*}

In this section, we elaborate on our insights in developing the Forecast \Recovery metrics in Section~\ref{sec:framework}. We first further discuss the shortcomings of existing evaluation metrics in CGA using real examples. Then, we discuss our insights into the relationship between our proposed \Recovery and the Accuracy metric. Finally, we present details of the exploratory empirical experiments (briefly discussed in Section~\ref{sec:framework}) to show that, unlike previous metrics, \Recovery metric effectively differentiates SOTA models from their naive counterparts, which lack the ability to process conversational context.
\subsection{Shortcomings of Existing Metrics}
Existing evaluation metrics for conversational forecasting, such as accuracy and F1 score, are largely adapted from standard classification tasks (with the notable exception of the mean horizon).
However, these metrics fail to account for the dynamic and online nature of the task, where forecasts are made incrementally as the conversation unfolds---one forecast for each new utterance.
Consequently, they cannot distinguish between a model that initially incorrectly predicts that the conversation will derail, but then \textit{recovers} by correcting its prediction after new utterances are observed (Model \textbf{\textcolor{blue}{B}} in the top example from Table \ref{tab:recovery-example}), and one that disregards new information and remains fixed on its initial misprediction (Model \textbf{\textcolor{red}{A}}).
Such forecast \recoveries are desirable both when the actual conversation recovers due to the tension dissipating and when the model's initial forecast was incorrect to start with due to an incorrect estimation.
%


\subsection{Relation Between \Recovery and Accuracy}
In Section~\ref{sec:framework}, we define  
\[
\mathrm{Recovery} = \frac{\corrcount}{N} - \frac{\incorrcount}{N},
\]
where $N$ is the size of the test set. In this section, we show that this metric is equivalent to the difference between the accuracy (in the current forecasting-formulation of this task, Section \ref{sec:related}), and the accuracy for a (hypothetical) classification-formulation of the task, where only the prediction at the end of the conversation ($\widehat{g_{n-1}}$) is evaluated.
(Recall that we have discarded the $n$-th utterance in the conversation where the event might occur.)

To establish the relationship between forecast \recovery and accuracy, we consider the cases in which \recovery occurs: $\widehat{g_k} = 1$ for some $k < n-1$ and $\binarylastforecast = 0$ (as defined in Section~\ref{sec:framework}).
%
%
There are two cases (1) the conversation eventually derails and thus the \recovery is incorrect (denoted as $\incorrcount$): this example adds to the forecasting-formulation accuracy and detracts from the classification-formulation accuracy; or (2) the conversation does not derail, and thus the \recovery is correct (denoted as $\corrcount$): this example detracts from the forecasting-formulation accuracy and adds to the classification-formulation accuracy. Therefore, counting all samples that are correctly predicted by both the conversation-level prediction ($\widehat{g_k}$) and the last-utterance prediction ($\widehat{g_{n-1}}$) is correct, we get:
\begin{equation*}
    \#\mathrm{Corr}(\widehat{g_k}) - \incorrcount = \#\mathrm{Corr}(\widehat{g_{n-1}}) - \corrcount
\end{equation*}
where $\#\mathrm{Corr}(\widehat{g_k})$ and $\#\mathrm{Corr}(\widehat{g_{n-1}})$ represent the number of samples where the $\widehat{g_k}$ is \textit{correct} and the number of samples where the $\widehat{g_{n-1}}$ is \textit{correct}, respectively. By further normalizing by the number of samples $N$ we can derive the following:
\begin{align*}
    \mathrm{Recovery} &= \frac{\corrcount - \incorrcount}{N} \\
    &= \#\frac{\mathrm{Corr}{(\widehat{g_{n-1}})}}{N} - \#\frac{\mathrm{Corr}(\widehat{g_k})}{N} 
\end{align*}
\noindent which is exactly the difference between the forecasting formulation accuracy and the classification formulation accuracy.

\subsection{Validation of Empirical Utility} 

Prior work has pointed out the ability to capture conversational context and the dynamics within as one of the main desired properties of conversational forecast models \cite{chang-danescu-niculescu-mizil-2019-trouble}.
Thus, to validate the Forecast \Recovery metric, we check whether it can differentiate between models that have this ability and those that do not.
To this end, we strip the best decoder-based generative model (Gemma2 9B) and best encoder-based model (DebERTaV3-large) of their access to conversational context and compare them with the full versions using the \Recovery metric and the existing metrics.
We find that, perhaps surprisingly for the DeBERTaV3 model, only the \recovery metric can clearly distinguish between the context-stripped and full versions, emphasizing the importance of considering this additional evaluation metric.

\xhdr{No-Context Setting}
We create the naive variant of the best-performing models by modifying the inputs while keeping the model architecture and parameters unchanged. 
Instead of providing the entire conversation context, $\mathrm{conversation}(t)$, to the model, we feed only the most recent utterance at each timestamp of the conversation, thus effectively eliminating the model's ability to consider the context in which each utterance appears.
Formally, in the normal setting, the individual forecast at timestamp $t$ is modeled as $    \indiforecast = \mathcal{P}(\mathrm{event}\,\big|\,\mathrm{conversation}(t))
$, where $\mathrm{conversation}(t)$ represents the full history of the conversation up to timestamp $t$. 
In contrast, under the No-Context setting, the probability is modeled as $$    \indiforecast = \mathcal{P}(\mathrm{event}\,\big|\,\mathrm{utterance}(t)),$$ where $\mathrm{utterance}(t)$ is the most recent utterance at timestamp $t$. 
%
%
Note that how we aggregate individual forecasts for the conversation-level forecast remains unchanged.

\begin{table}[!ht]
\resizebox{\linewidth}{!}{%
\begin{tabular}{lccl}
\toprule
Model             & Acc$\uparrow$ & F1$\uparrow$ & Recovery$\uparrow$ \\
\phantom{Mo} w/ context? &               &              & ($CA/N-IA/N$) \\
\midrule
DeBERTaV3~ Yes & $68.9$ & $70.3$ & $+1.1$ ($7.6 - 6.5$) \\
\phantom{DeBERTaV3~~} No & $67.9$ & $70.7$ & $-3.5$ $(13.4 - 16.9)$ \\
\midrule
Gemma2~~~~~~~~ Yes & $71.0$ & $72.3$ & $+1.8$ ($8.4 - 6.6$) \\
\phantom{Gemma2~~~~~~~~} No & $68.7$ & $68.8$ & $-4.7$ $(10.0 - 14.7)$ \\
\bottomrule
\end{tabular}%
}
\caption{Comparing the performance of DeBERTaV3-large and Gemma2 9B on \newcmv with (Normal) and without (No-Context) access to conversational context.
}
\label{tab:normal-nocontext}
\end{table}

\xhdr{Empirical Comparison}
The results of our experiments are presented in Table~\ref{tab:normal-nocontext}.
Surprisingly, our results show that the absence of conversational context has a relatively small impact on DeBERTaV3's performance under the main evaluation metrics (accuracy and F1).
On the other hand, we observe a substantial difference in $\mathrm{\Recovery}$: DeBERTaV3 has a positive score ($+1.1$), indicating that the model maintains a positive balance between correct and incorrect \recoveries, while No-Context DeBERTaV3 has a negative score ($-3.5$), suggesting that it produces significantly more incorrect than correct \recoveries.
In the case of Gemma2, the accuracy metric provides only a small difference ($2.3$) between the full and No-Context variants. In contrast, \Recovery offers a clearer difference ($6.5$), revealing that the lack of context leads to a significantly higher number of incorrect \recoveries.
\section{The Importance of Threshold Tuning}
\label{appendix:t-tuning}
Threshold tuning involves learning the decision threshold $T$ (in $\widehat{g_t} = (\indiforecast>T)$) to maximize the performance of models on the development split of the dataset. In this work, we carefully perform threshold tuning following the practice outlined by \citet{chang-danescu-niculescu-mizil-2019-trouble}. However, this practice has not been mentioned or discussed by subsequent studies \cite{kementchedjhieva-sogaard-2021-dynamic,yuan_conversation_2023}. We hypothesize that the lack of consistency on whether to tune the decision threshold on the development split before benchmarking performance on the test split may lead to significant discrepancies in performance reports across existing studies for the same model.

\xhdr{Ablation Study}
In this section, we conduct an ablation study to investigate the effect of threshold $T$ tuning on the performance of CGA models. 

\begin{table}[!ht]
\centering
\begin{tabular}{l|cc}
\hline
\multirow{2}{*}{Model} & \multicolumn{2}{c}{Acc $\uparrow$ }                                                                            \\ \cline{2-3} 
                       & \multicolumn{1}{c|}{$T$ tuning} & \begin{tabular}[c]{@{}c@{}}\textbf{No} \\ $T$ tuning\end{tabular} \\ \hline
DeBERTaV3-large        & \multicolumn{1}{c|}{$68.9$}           & $67.4$                                                         \\
RoBERTa-large          & \multicolumn{1}{c|}{$68.6$}           & $66.8$                                                        \\ 
Gemma2 9B              & \multicolumn{1}{c|}{$71.0$}           & $67.4$                                                         \\
LlaMa3.1 8B            & \multicolumn{1}{c|}{$70.0$}           & $66.5$                                                         \\
Mistral 7B             & \multicolumn{1}{c|}{$70.7$}           & $68.7$                                                         \\
Phi4 14B               & \multicolumn{1}{c|}{$70.5$}           &      $69.2$                                                        \\
\hline
\end{tabular}
\caption{Ablation study on the effect of threshold tuning on the performance of the top 6 CGA models.}
\label{tab:ablationthreshtuning}
\end{table}

Table~\ref{tab:ablationthreshtuning} illustrates the impact of threshold tuning on the performance of the top 6 CGA models. 
Unlike many other NLP tasks---where threshold tuning is rarely a critical factor in model performance---we observe that the performance of these 6 SOTA models drops markedly, by approximately 1.3 (Phi4 14B) to 3.6 (Gemma2 9B), without threshold tuning. This result highlights the importance of threshold tuning in conversational forecasting tasks (such as the CGA task).

\xhdr{Explanation}
We believe this phenomenon stems from a fundamental difference between classification and forecasting tasks (see Section~\ref{sec:related}). 
While classification tasks evaluate each prediction against a ground-truth label, conversational forecasting aggregates multiple predictions into a single conversation-level forecast before evaluation.
However, this difference is not reflected in how models are trained for forecasting. 

\begin{figure}[!ht]
    \centering
    \includegraphics[width=\linewidth]{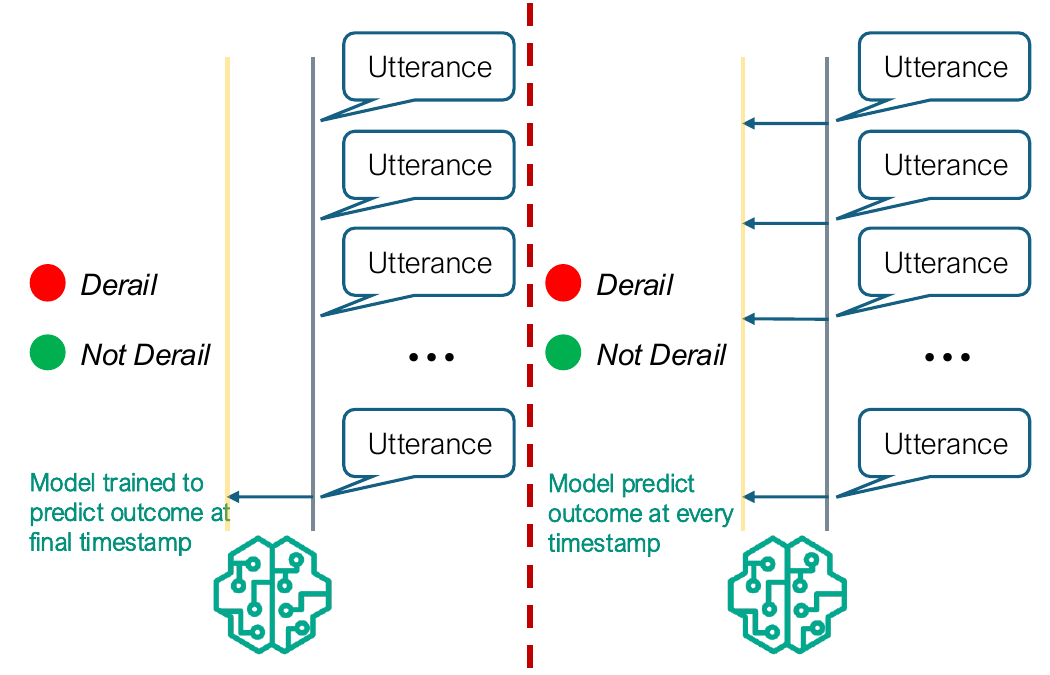}
    \caption{
        \textbf{Illustration of Training (left) and Testing (right) CGA models.}
        During evaluation, models must generate forecasts for all timestamps in a conversation.
        However, models are trained only to distinguish between derailing and non-derailing conversations at the last timestamp $(t = n - 1)$ of each conversation.
    }
    \label{fig:train_test_illustration}
\end{figure}

As described in Section~\ref{sec:framework}, the most effective training strategy is to use snapshots of the conversation taken at the last timestamp before the final utterance (illustrated in Figure~\ref{fig:train_test_illustration}). 
From our perspective, this approach underutilizes the training data, especially in conversations that nearly derail but ultimately remain civil. 
For example, a conversation might include several hostile or impolite exchanges before the participants try to resolve the conflict and recover the conversation. 
When models are trained only on the final snapshot of such conversations, they may learn only the outcome---recovery---while missing the ``hard moments'' that could resemble actual derailments. 
As a result, at test time, models may struggle to distinguish between ``hard moments'' that do and do not lead to derailment.

This training-inference mismatch may explain why threshold tuning is critical in conversational forecasting. 
We hypothesize that the optimal threshold $T$ often needs to be greater than $0.5$ to adjust for the model's overestimation of derailment risk in ``hard moments.''
To validate this, we examine the 30 optimal thresholds $T$ identified in the 6 best performing CGA models in Table~\ref{tab:survey-cmv_large}, and find that 28 out of 30 exceed $0.5$. 
The mean of these thresholds is $0.65$, with a standard deviation of $0.08$. 
A one-sided t-test further confirms that the mean is significantly greater than $0.5$ ($p = 6.9 \times 10^{-11}$).

\section{Details for Models Implementation}
\label{appendix:modelimplementation}
\subsection{Decoder-based generative models}
\renewcommand{\arraystretch}{1.3}
\rowcolors{2}{gray!10}{white}

\begin{table*}[!ht]
\centering
\begin{tabular}{{l|c|cc|c}}
\toprule
\multirow{1}{*}{Hyperparameter} & \multirow{1}{*}{CRAFT} & \multicolumn{2}{c|}{Encoder-based Models} & \multirow{1}{*}{Decoder-based Models} \\
                &                        & \multicolumn{1}{c}{base}     & large    &                                    \\ 
\midrule
Train batch size                        & $64$         & $4$            & $12$           & $32$ \\
Max sequence length                     & $/$          & $512$          & $512$          & $8{,}192$ \\
LR scheduler                            & Linear       & Linear         & Linear         & Linear \\
Optimizer                               & Adam         & AdamW          & AdamW          & AdamW \\
Number of epochs                        & $10$         & $4$            & $4$            & $4$ \\
Learning rate                           & $1\text{e}{-5}$ & $6.7\text{e}{-5}$ & $6.7\text{e}{-5}$ & $1\text{e}{-4}$ \\
NVIDIA GPU used                         & RTX $2080$   & RTX $A6000$      & RTX $A6000$      & RTX $A6000$ \\
\makecell[l]{Training time\\(on \texttt{\newcmv})} & 30 mins     & 1 hour         & 2.5 hours      & 10 hours \\
\bottomrule
\end{tabular}
\caption{Hyperparameters used for training or fine-tuning different models.}
\label{table:hyperparameters}
\end{table*}

\rowcolors{2}{}{}

To fine-tune decoder-based generative models (e.g., Gemma, LLaMA, Mistral, and Phi) on CGA datasets, we use the following prompt:
\subsubsection*{Prompt for CGA tasks}


\begin{tcolorbox}[colback=gray!5, colframe=black!50, boxrule=0.5pt, arc=4pt, left=5pt, right=5pt, top=5pt, bottom=5pt]
\textbf{Instruction:}

You are a moderator observing an ongoing conversation. Your goal is to determine whether the conversation will derail into a personal attack.

Pay attention to conversational flow and speaker dynamics. Be careful—sensitive topics do not always lead to personal attacks.

\vspace{1em}
\textbf{Conversation Transcript:}
\begin{quote}
\texttt{\{Conversation\_transcript\}}
\end{quote}

Will the above conversation derail into a personal attack now or at any point in the future?

\textbf{Strictly start your answer with} \texttt{Yes} \textbf{or} \texttt{No}; any other format is invalid.
\end{tcolorbox}

We then use the logits of the first generated token, constrained to the two vocabulary options ``Yes'' and ``No,'' to compute the individual forecasts from the models. 
The individual forecast can be formally written as:
$$\hat{y_t} = \frac{e^{yes\_logit}}{e^{yes\_logit} + e^{no\_logit}},$$
where $yes\_logit$ and $no\_logit$ are scalar outputs from generative models.

Due to computational limitations, we use a LoRA rank of 64 to fine-tune 4-bit quantized models, contributed by Unsloth.\footnote{https://unsloth.ai}

\subsection{Encoder-based models}
To fine-tune Encoder-based models (BERT, RoBERTa, SpanBERT, and DeBERTaV3) on CGA datasets, we represent the input sequence as 

\texttt{\textcolor{blue}{[CLS]}~Utt\textsubscript{1}~\textcolor{red}{[SEP]}~Utt\textsubscript{2}~\textcolor{red}{[SEP]}~$\ldots$~Utt\textsubscript{n}~\textcolor{red}{[SEP]}},

\noindent where \texttt{\textcolor{blue}{[CLS]}} and \texttt{\textcolor{red}{[SEP]}} are special tokens defined by the tokenizers associated with these models.
If the length of the tokenized sequence exceeds 512---the maximum sequence length supported by our model---we truncate tokens from the left side of the sequence.
We then use the hidden state of the last layer corresponding to the first input token (\texttt{\textcolor{blue}{[CLS]}}) as the aggregate representation and introduce new classification layer weights during fine-tuning to classify/forecast from the aggregate representation. 
We use cross-entropy loss during the fine-tuning process.

\section{Details for Models Training}
\label{appendix:modeltraining}

In this section, we provide detailed information on the software and hardware used for training the models to ensure the reproducibility of the results reported in this paper.

Regarding software dependencies, we list their licenses below:
\begin{enumerate} \item \textbf{PyTorch} \cite{10.5555/3454287.3455008}, a deep learning framework, is licensed under BSD-3.
\item \textbf{Transformers} library \cite{wolf-etal-2020-transformers}, used for model training and evaluation, is licensed under the Apache License, Version 2.0.
\item The \textbf{Unsloth} library\footnote{https://docs.unsloth.ai}, used for quantization and low-rank adaptation of decoder-based generative models, is licensed under the Apache License, Version 2.0.
\item \textbf{ConvoKit} \cite{chang-etal-2020-convokit}, used to develop our unified evaluation framework (Section~\ref{sec:framework}), is licensed under the MIT License. \end{enumerate}

Table~\ref{table:hyperparameters} summarizes the hyperparameters and GPU resources used for training/fine-tuning our models.

\end{document}